\documentclass[sigconf]{acmart}
\AtBeginDocument{%
  }
\usepackage{multirow}
\usepackage[table]{xcolor}
\usepackage[dvipsnames]{xcolor}
\usepackage{url}
\usepackage{indentfirst}
\usepackage{tabularx}
\usepackage{hyperref}
\definecolor{clip4geopink}{RGB}{239,151,162}
\copyrightyear{2025}
\acmYear{2025}
\setcopyright{cc}
\setcctype{by}
\acmConference[GeoAI '25]{The 8th ACM SIGSPATIAL International Workshop on AI for Geographic Knowledge Discovery}{November 3--6, 2025}{Minneapolis, MN, USA}
\acmBooktitle{The 8th ACM SIGSPATIAL International Workshop on AI for Geographic Knowledge Discovery (GeoAI '25), November 3--6, 2025, Minneapolis, MN, USA}
\acmDOI{10.1145/3764912.3770830}
\acmISBN{979-8-4007-2179-3/2025/11}

\begin{document}

%\title{Semantic4Safety: Causal Inference on Street View Imagery for Urban Road Safety}
\title{Semantic4Safety: Causal Insights from Zero-shot Street View Imagery Segmentation for Urban Road Safety}

\renewcommand{\shorttitle}{Semantic4Safety}

\author{
Huan Chen$^{1, \#}$,
Ting Han$^{1,2, \#}$,
Siyu Chen$^{2}$,
Zhihao Guo$^{3}$,
Yiping Chen$^{1,*}$,
Meiliu Wu$^{2,*}$
}

\affiliation{
    \city{$^{1}$ School of Geospatial Engineering and Science, Sun Yat-sen University, Zhuhai, China} \\
    \institution{$^{2}$ School of Geographical and Earth Sciences, University of Glasgow, Glasgow, United Kingdom}
    \country{$^{3}$ School of Economics and Management, Shanxi University, Taiyuan, China}
}

\begin{abstract}
Street-view imagery (SVI) offers a fine-grained lens on traffic risk, yet two fundamental challenges persist: (1) \textit{how to construct street-level indicators that capture accident-related features}, and (2) \textit{how to quantify their causal impacts across different accident types}. To address these challenges, we propose \textbf{Semantic4Safety}, a framework that applies zero-shot semantic segmentation to SVIs to derive 11 interpretable streetscape indicators, and integrates road type as contextual information to analyze approximately 30,000 accident records in Austin. Specifically, we train an eXtreme Gradient Boosting (XGBoost) multi-class classifier and use Shapley Additive Explanations (SHAP) to interpret both global and local feature contributions, and then apply Generalized Propensity Score (GPS) weighting and Average Treatment Effect (ATE) estimation to control confounding and quantify causal effects. Results uncover heterogeneous, accident-type-specific causal patterns: features capturing scene complexity, exposure, and roadway geometry dominate predictive power; larger drivable area and emergency space reduce risk, whereas excessive visual openness can increase it. By bridging predictive modeling with causal inference, Semantic4Safety supports targeted interventions and high-risk corridor diagnosis, offering a scalable, data-informed tool  for urban road safety planning. 
\end{abstract}

%%
%% The code below is generated by the tool at http://dl.acm.org/ccs.cfm.
%% Please copy and paste the code instead of the example below.
%%


%%
%% Keywords. The author(s) should pick words that accurately describe
%% the work being presented. Separate the keywords with commas.
%% A "teaser" image appears between the author and affiliation
%% information and the body of the document, and typically spans the
%% page.
\begin{teaserfigure}
    \setlength{\abovecaptionskip}{0.1cm}
    \setlength{\belowcaptionskip}{-0.1cm}
  \includegraphics[width=\textwidth]{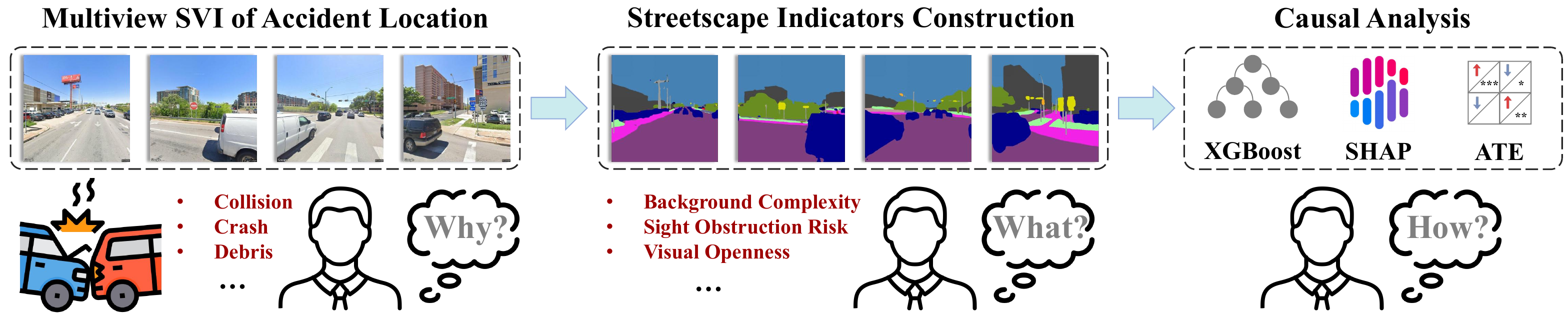}
  \caption{\textbf{Overview of our proposed \textit{Semantic4Safety} framework.} 
    The framework follows a \textit{Why–What–How} logic to analyze traffic accidents using street-view imagery. (1) We start from approximately 30,000 real-world accident locations, collecting over 120,000 multi-view street-view images covering diverse urban contexts to ask: \textit{Why do these accidents happen?} (2) Through zero-shot semantic segmentation, we construct a set of interpretable streetscape indicators to reveal: \textit{What factors contribute to accident occurrence?} (3) Finally, we employ XGBoost multi-class modeling, SHAP interpretability, and causal effect estimation to uncover: \textit{How do these factors causally influence different accident types?} This structured progression bridges observation, feature construction, and causal inference to provide both explanatory and actionable insights for urban road safety.}
  \label{fig: overview}
  \vspace{0.4cm}
\end{teaserfigure}

  % \caption{Our proposed framework \textit{Semantic4Safety} for traffic accident analysis based on street-view imagery zero-shot semantic segmentation and advanced analytical methods. The framework consists of three main stages: (1) \textit{Multiview SVI of Accident Location.} We collected over 120,000 street-view images from approximately 30,000 real-world traffic accident locations, covering diverse urban scenes with multi-angle, multi-time, and multi-environment perspectives. (2) \textit{Street-view Indicators Construction.} We employ zero-shot semantic segmentation to perform pixel-level analysis on the collected SVI. From this, we construct 12 quantitative street-scene indicators, encompassing key features related to spatial structure, visual complexity, visibility risk, and road configuration. (3) \textit{Causal Analysis.} We integrate XGBoost multiclass classification, SHAP-based interpretability, and Average Treatment Effect estimation to identify causal relationships between street-view indicators and various types of traffic accidents.}

%%
%% This command processes the author and affiliation and title
%% information and builds the first part of the formatted document.
\maketitle
%\footnotetext[1]{$\dagger$ These authors contributed equally to this work.}
%\footnotetext[2]{$\ast$ Corresponding authors.}

\section{Introduction}
In recent years, deep learning–based artificial intelligence technologies have injected new momentum into urban traffic safety research, fostering smarter traffic management, enhanced public safety, and sustainable urban development \cite{irfan2024toward,yang2025theory}. In particular, SVI has emerged as a critical tool for analyzing traffic accident risks. Its high resolution and extensive coverage offer unique advantages for extracting dynamic streetscape features \cite{biljecki2021street,Rzotkiewicz_Pearson_Dougherty_Shortridge_Wilson_2018,Kang_Zhang_Gao_Lin_Liu_2020}. Moreover, by conducting in-depth mining of SVI and applying task-specific fine-tuning \cite{Shu_Yan_Xu_2021,Arya_Maeda_Ghosh_Toshniwal_Mraz_Kashiyama_Sekimoto_2021}, researchers can more accurately uncover the relationship between the traffic environment and accidents, opening a new phase of traffic safety analysis centered on SVI \cite{Jiao_Wang_2023}.

Data-driven traffic accident analysis models focus on utilizing SVI to identify spatial and temporal patterns of traffic accidents and to enhance the understanding of the relationship between environmental features and accidents \cite{ijgi12020073}. Semantic segmentation is applied to extract visual features from street-level images for accident analysis \cite{5459249,9724728,Dai02062024}, and causal inference methods are used to uncover the causal relationships between these features and accident risks. Furthermore, SVI-based analysis models demonstrate strong generalization capabilities and are a key focus for advancing knowledge transfer in traffic safety research.

However, existing semantic features typically rely only on generic annotations and lack integration with structured traffic contexts such as accident type or spatial distribution \cite{guo2024fusion,ye2025street}. This limits the capacity of models to capture the nuanced dynamics and spatial heterogeneity of traffic environments \cite{cai2022applying}. Although prior studies have explored combining visual features with geospatial coordinates, location information alone often fails to reflect the underlying complexity of street-level accident risk \cite{rui2016network,castro2015spatial}. In effect, two critical gaps remain: the lack of a unified framework that aligns visual semantics with structured accident data, and the persistent inability of conventional models to extract and utilize risk-relevant features in dynamic, real-world street scenes \cite{hu2023uncovering}. These limitations motivate us to develop a new framework that explicitly integrates SVI with structured indicators to enhance causal understanding of traffic accidents and to support scalable, transferable safety interventions.

This paper presents a novel traffic accident analysis framework, named \textbf{Semantic4Safety}, designed to address the above two major challenges. The framework is built upon a large-scale dataset comprising 120,000 street-view images collected from approximately 30,000 accident locations in Austin, Texas, and leverages zero-shot semantic segmentation to extract 11 interpretable streetscape indicators. These indicators span multiple dimensions, including sight obstruction risk, vegetation coverage, traffic sign completeness, drivable area ratio, and building occlusion, and are designed to capture visual-spatial elements associated with accident risk. 

To analyze their impact, we apply a multi-stage causal inference pipeline. Specifically, we use XGBoost for multi-class accident prediction, Shapley Additive Explanations (SHAP) to interpret both global and local feature importance, and Generalized Propensity Score (GPS) weighting to adjust for confounders in causal estimation. We then estimate Average Treatment Effects (ATE) and construct a causal effect matrix, which reveals heterogeneous causal impacts of different streetscape indicators on distinct accident types. Experimental results demonstrate that Semantic4Safety achieves robust performance in both feature modeling and causal analysis. The framework also supports interpretable risk assessment and targeted intervention strategy development, offering practical utility for urban traffic safety planning and scalable deployment in diverse geospatial contexts. The main contributions of this study are summarized as follows:
\begin{itemize}
    \item We propose a novel framework that applies zero-shot semantic segmentation to SVI for traffic accident analysis, constructing 11 interpretable streetscape indicators from 30,000 accident cases in Austin.
    \item We design an efficient evaluation pipeline that leverages XGBoost and multi-source data to assess the predictive power of key visual and contextual features, enabling robust causal analysis.
    \item We integrate Generalized Propensity Score weighting and ATE estimation to quantify indicator-specific causal effects across five accident types, producing a fine-grained, statistically supported causal effect matrix.
\end{itemize}
\section{Related Work}

\subsection{SVI Analysis with Semantic Segmentation}

Street-view imagery (SVI) has become a vital resource in urban research due to its fine-grained visual detail and broad geographic coverage \cite{RUI2023104472,FAN2024105862,CHEN2023104329}. Compared to traditional remote sensing, SVI excels at capturing streetscape elements that are otherwise difficult to observe \cite{CUI2023103537,Cao_Zhu_Tu_Li_Cao_Liu_Zhang_Qiu_2018,Fan_Zhang_Loo_Ratti}. Semantic segmentation further enhances the value of SVI by automatically partitioning images into meaningful urban categories, such as roads, sidewalks, vegetation, and vehicles. Unlike earlier methods based on low-level color or texture cues (e.g., K-means clustering \cite{8ddb7f85-9a8c-3829-b04e-0476a67eb0fd}, region growing \cite{295913}), modern deep learning–based segmentation \cite{chen2024depth,chen2025hspformer,han2024epurate} enables more accurate, robust, and consistent extraction of urban features across complex environments. This synergy between SVI and semantic segmentation has opened new avenues for analyzing urban form \cite{Li_Zhang_Li_Ricard_Meng_Zhang_2015,Liang_Gong_Sun_Zhou_Li_Li_Liu_Shen_2017}, environmental quality \cite{Yang_Zhao_Mcbride_Gong_2009}, and traffic safety \cite{Koo_Guhathakurta_Botchwey_2022}.

\subsection{Causal Inference in Traffic Safety}

Causal inference provides a critical complement to correlation-based prediction in traffic safety research \cite{DAVIS200095,Ma01112024}. Despite the widespread use of machine learning models \cite{Breiman_2001,Hearst_Dumais_Osuna_Platt_Scholkopf_1998} for crash prediction, most approaches remain associative in nature. Post-hoc feature attributions or black-box explanations offer limited practical guidance, and counterfactual questions, such as how crash risk would change if a streetscape element were altered—are rarely explored.
Furthermore, issues such as confounding, imbalance, and spatial correlation are often overlooked, with limited use of propensity-based balancing methods \cite{ROSENBAUM_RUBIN_1983}. As a result, findings may lack actionable value and struggle to generalize across diverse urban contexts. Integrating causal inference methods into traffic safety analysis enables estimation of the actual effect of specific features on crash occurrence and severity, providing more robust and transferable evidence for urban design, policymaking, and risk mitigation \cite{10.1145/3444944}.

Compared to existing studies, our paper introduces two key innovations: First, we leverage zero-shot semantic segmentation to construct a high-resolution, generalizable indicator system from multi-view SVI, enabling structured and interpretable analysis of fine-grained urban morphology. Second, we integrate causal inference methods (GPS weighting and ATE estimation) to move beyond correlation and quantify the causal impact of street-level features on multiple accident types, offering actionable insights for safety-oriented urban design and intervention.
\section{Methodology}

\begin{figure*}[!t]
    \setlength{\abovecaptionskip}{0.1cm}
    \setlength{\belowcaptionskip}{-0.1cm}
    \centering
    \includegraphics[width=\textwidth]{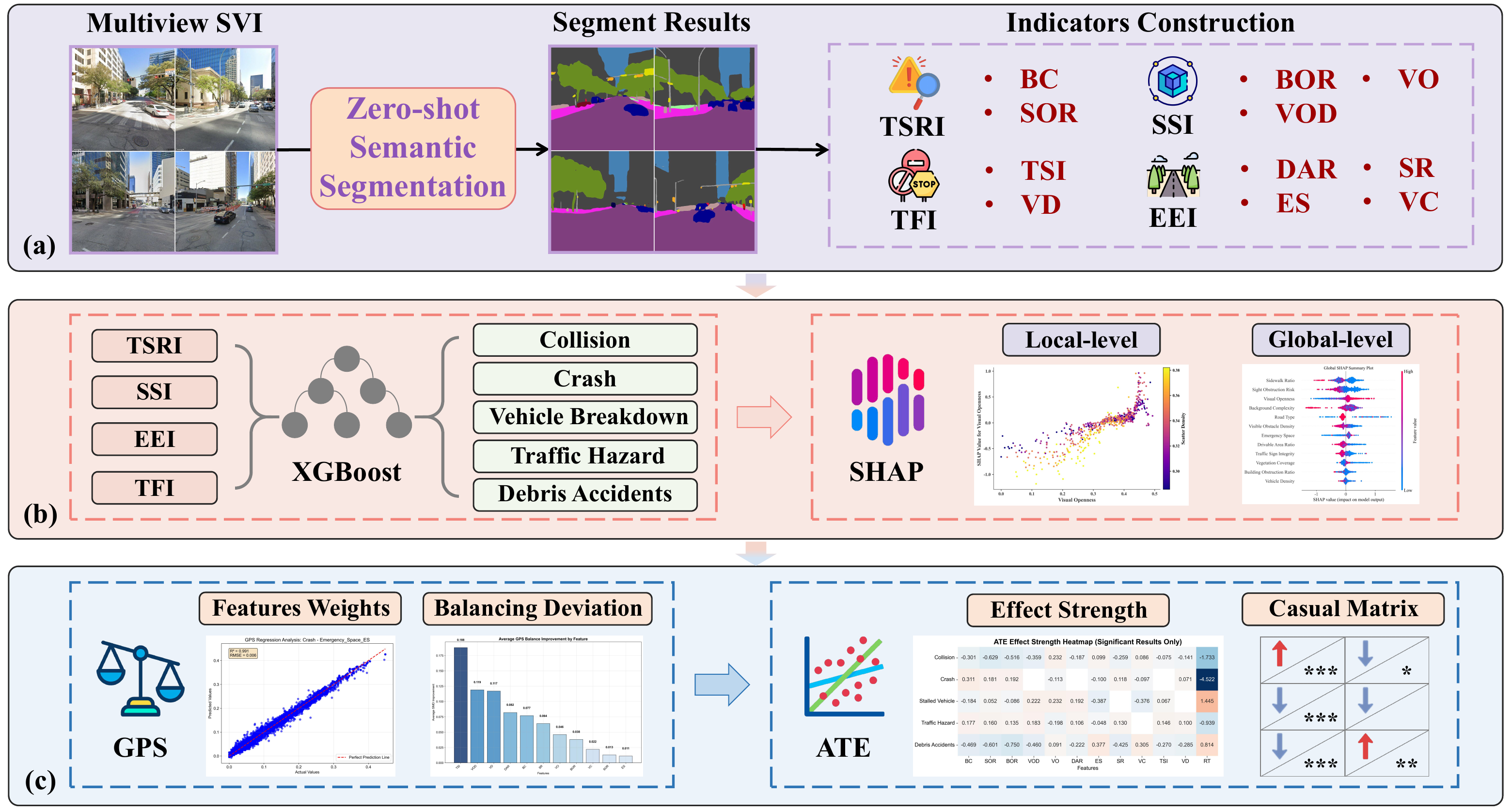}
    \caption{\textbf{Overview of the proposed Semantic4Safety framework.} We first process SVI using zero-shot segmentation to extract 11 indicators across four categories. These indicators are then used to predict five accident types through XGBoost, with SHAP providing both local and global-level interpretability. Finally, Generalized Propensity Score (GPS) weighting and ATE estimation are applied to quantify the causal effects of indicators across accident categories.}
%(a) \textit{SVI Semantic Segmentation.} Multiview SVI are segmented using zero-shot semantic models to calculate diverse streetscape indicators, including Traffic Safety Risk Indicators (TSRI), Spatial Structure Indicators (SSI), Environmental Element Indicators (EEI), and Traffic Facility Indicatorss (TFI). (b) \textit{XGBoost Multi-class Modeling and SHAP Interpretability Analysis.} Street-view indicators are used to predict five accident types through XGBoost, with SHAP providing both local-level and global-level interpretability of feature contributions. (c) \textit{Causal Inference Analysis.} GPS weighting is applied to control for confounding, followed by ATE estimation and construction of the causal effect matrix, quantifying the causal influence of streetscape indicators on different accident risks.}
    \label{fig: framework}
\end{figure*}

We propose a framework named Semantic4Safety for traffic accident analysis based on SVI and advanced analytical methods, as shown in Fig.~\ref{fig: framework}. The framework systematically processes large-scale accident-related SVI through the following modules: (a) We collect multiview SVI from 30,000 accident locations and apply a zero-shot semantic segmentation to generate pixel-level semantic information. From these outputs, 11 indicators are derived from SVI plus road type (12 features total) to capture both visual cues and spatial context. (b) Leveraging the constructed indicators and corresponding accident labels, we use XGBoost and SHAP to interpret the predictions at both global and local levels, identifying the most influential street features for each accident category. (c) To move beyond correlational analysis, we integrate GPS weighting with ATE estimation. This allows us to quantify the causal effect strength of each indicator on different accident types, generating a causal effect matrix that reveals heterogeneity in risk contributions.

\subsection{Data Collection}

\begin{figure}[!t]
    \setlength{\abovecaptionskip}{0.1cm}
    \setlength{\belowcaptionskip}{-0.1cm}
    \centering
    \includegraphics[width=1\linewidth]{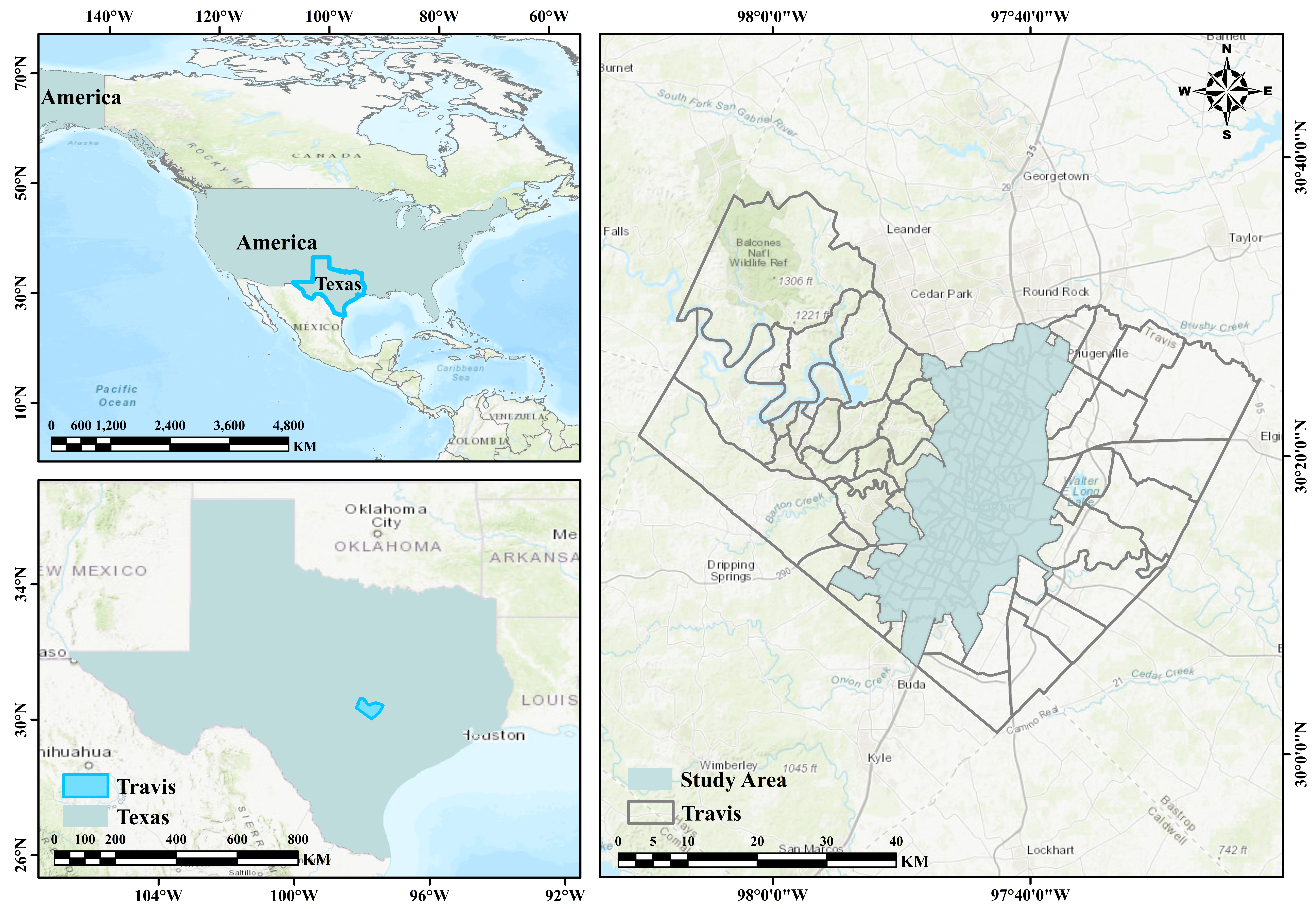}
    \caption{The study area is located in Austin, Texas, United States. The main panel provides a detailed view of the study area boundaries within Austin, where SVI and traffic accident data were collected and analyzed.}
    \label{fig: Study_Area}
    \vspace{-0.3cm}
\end{figure}

As shown in Fig.~\ref{fig: Study_Area}, we select the main area in Austin, Texas, as the study area based on the openness, integrity, and high quality of the available data. The data collection encompasses three core modalities: textual accident records, geospatial data, and SVI.

\paragraph{Textual Records.} We obtain real-time traffic accident reports from the City of Austin Open Data Portal\footnote{https://data.austintexas.gov/}, sourced via the Combined Transportation, Emergency, and Communications Center. Each report includes the incident time, accident type, and precise geographic coordinates. A total of 18 accident categories are documented and reclassified by domain experts into five main types: Collision, Crash, Vehicle Breakdown, Traffic Hazard, and Debris Accidents. After filtering for records within the study area, we retain 31,983 valid accident cases from February 2024 to January 2025 for downstream analysis.

\paragraph{Geospatial Data.} Road network data for Austin are sourced from OpenStreetMap (OSM)\footnote{https://www.openstreetmap.org/}, and projected to the WGS84 coordinate system. To ensure spatial accuracy, geospatial experts manually corrected missing labels, coordinate misalignments, and ambiguous classifications\cite{balaban2022understanding}. The refined roads are finally reclassified into four categories based on road type, as summarized in Tab.~\ref{tab:road-classification}.

\paragraph{SVI} For each accident location, we retrieve the corresponding SVI from Google Maps\footnote{https://www.google.com/maps/} using its latitude and longitude. Four images are captured from different directions (0 degree, 90 degree, 180 degree, and 270 degree), each with a resolution of $ 640 \times 640 $ pixels.

\begin{table}[!t]
    \setlength{\abovecaptionskip}{0.1cm}
    \setlength{\belowcaptionskip}{-0.1cm}
    \centering
    \caption{Reclassification of OSM road types into four hierarchical categories.}
    \resizebox{0.48\textwidth}{!}{
    \begin{tabular}{ll}
    \toprule
    \textbf{Road Category} & \textbf{OSM Road Types} \\
    \midrule
    Path & footway, path, cycleway, pedestrian \\
    Linkroad & motorway link, trunk link, primary link \\
    Specialroad & service, track, unclassified, residential \\
    Principal tag & motorway, trunk, primary, secondary, tertiary \\
    \bottomrule
    \end{tabular} \label{tab:road-classification}}
\end{table}

\begin{table}[!t]
\setlength{\abovecaptionskip}{0.1cm}
\setlength{\belowcaptionskip}{-0.1cm}
\centering
\caption{Quantitative Definitions of Traffic Safety Indicators Derived from Semantic Segmentation of SVI.}
\resizebox{0.48\textwidth}{!}{
\begin{tabular}{lll}
\toprule
\textbf{Category} & \textbf{Indicators} & \textbf{Calculation Formula} \\
\midrule
\multirow{2}{*}{TSRI} & BC & $-\sum (p_i \cdot \log p_i) / \log N$ \\
& SOR & $\text{Pixels}_{\text{obstruction}} / \text{Pixels}_{\text{center region}}$ \\
\midrule
\multirow{3}{*}{SSI} & BOR & $\text{Pixels}_{\text{buildings}} / \text{Pixels}_{\text{total}}$ \\
& VOD & $\text{Connected components} / (\text{Image area} / 10000)$ \\
& VO & $(\text{Pixels}_{\text{sky}} + \text{Pixels}_{\text{terrain}}) / \text{Pixels}_{\text{total}}$ \\
\midrule
\multirow{4}{*}{EEI} & DAR & $\text{Pixels}_{\text{road}} / \text{Pixels}_{\text{total}}$ \\
& ES & $\text{Pixels}_{\text{escape space}} / \text{Pixels}_{\text{total}}$ \\
& SR & $\text{Pixels}_{\text{sidewalk}} / \text{Pixels}_{\text{total}}$ \\
& VC & $\text{Pixels}_{\text{vegetation}} / \text{Pixels}_{\text{total}}$ \\
\midrule
\multirow{2}{*}{TFI} & TSI & $\frac{1}{n}\sum_{i=1}^{n} \frac{4\pi \cdot \text{Area}_i}{\text{Perimeter}_i^2}$ \\
& VD & $\text{Pixels}_{\text{vehicles}} / \text{Pixels}_{\text{total}}$ \\
\bottomrule
\end{tabular}}\label{tab:safety_metrics}
\end{table}

\subsection{Zero-shot Semantic Segmentation and Indicator Construction}

To extract structured features from SVI, we adopt our previous work \textbf{Vireo}\cite{chen2025leveragingdepthlanguageopenvocabulary} to perform zero-shot semantic segmentation. Unlike conventional closed-set models, Vireo enables flexible and open-vocabulary parsing of SVI, allowing for the identification of diverse semantic categories beyond predefined taxonomies, which is an essential capability for urban traffic safety analysis.

Formally, given an SVI $I \in \mathbb{R}^{H \times W \times 3}$, Vireo encodes it into a high-dimensional representation using a vision VFM $f_{\theta}(\cdot)$:
\begin{equation}
F = f_{\theta}(I), \quad F \in \mathbb{R}^{H' \times W' \times d},
\end{equation}
where $F$ denotes the latent feature map, with $H'$ and $W'$ representing the downsampled dimensions and $d$ the embedding dimension.

To enable zero-shot segmentation, Vireo incorporates textual descriptions of arbitrary categories into the segmentation pipeline. Each class label $c \in \mathcal{C}$ is embedded by a LLM encoder $g_{\phi}(\cdot)$:
\begin{equation}
t_c = g_{\phi}(c), \quad t_c \in \mathbb{R}^d.
\end{equation}
The LLM encoder is contrastively aligned with the CLIP-style pretrained backbone, ensuring that image and text embeddings lie in a shared feature space.

Vireo computes the similarity between each pixel feature and the text embedding of a given class, to generate segmentation logits:
\begin{equation}
S(x,y,c) = \langle F(x,y), t_c \rangle,
\end{equation}
where $S(x,y,c)$ denotes the compatibility score between pixel $(x,y)$ and category $c$. Leveraging the joint embedding space, Vireo recognizes both traffic-specific objects (e.g., vehicles, traffic signs) and broader urban elements (e.g., trees, buildings) without task-specific retraining.

A softmax is then applied over all candidate categories to obtain pixel-wise probabilities:
\begin{equation}
P(x,y,c) = \frac{\exp(S(x,y,c))}{\sum_{c' \in \mathcal{C}} \exp(S(x,y,c'))}.
\end{equation}
Finally, the segmentation mask $M \in \mathbb{R}^{H \times W}$ is generated by assigning each pixel to the category with the highest probability:
\begin{equation}
M(x,y) = \arg\max_{c \in \mathcal{C}} P(x,y,c).
\end{equation}

Compared to traditional fixed-class segmentation models, Vireo offers two key advantages: (1) \textit{Scalability}: it supports the inclusion of new traffic-related categories simply by providing their textual descriptions, and (2) \textit{Adaptability}: it generalizes well to diverse urban scenes beyond the training distribution. Therefore, we do not require fine-grained semantic annotations for Austin’s SVI. Instead, we can directly extract structured traffic safety indicators, such as sidewalk ratio, vegetation coverage, and visibility constraints. The resulting semantic masks form the critical foundation for subsequent quantitative analysis in our framework.

We derive eleven traffic safety–related indicators from the segmentation masks using pixel counting and basic morphological operations. These indicators are grouped into four categories:
\begin{itemize}
    \item Traffic Safety Risk Indicators (TSRI): Background Complexity (BC), Sight Obstruction Risk (SOR).
    \item Spatial Structure Indicators (SSI): Building Obstruction Ratio (BOR), Visible Obstacle Density (VOD), Visual Openness (VO).
    \item Environmental Element Indicators (EEI): Drivable Area Ratio (DAR), Emergency Space (ES), Sidewalk Ratio (SR), Vegetation Coverage (VC).
    \item Traffic Facility Indicators (TFI): Traffic Sign Integrity (TSI), Vehicle Density (VD).
\end{itemize}

Streetscape–derived traffic safety indicators and their computational formulations are shown in Tab.~\ref{tab:safety_metrics}. Each indicator is first computed per view, aggregated at the point level, and averaged across four cardinal directions. The resulting features are then merged with accident records and road-type data to form a unified dataset used for multi-class modeling, SHAP interpretability analysis, and causal inference.

\subsection{XGBoost Multi-class Modeling with SHAP Interpretability Analysis}

XGBoost is a scalable gradient boosting framework that ensembles decision trees with explicit regularization. Designed for large-scale, heterogeneous tabular data, it offers fast training and strong generalization—making it well-suited for modeling streetscape features at scale.

In our framework, accident type classification is formulated as a five-class problem based on domain-informed aggregation. Each input sample includes 12 features: 11 streetscape indicators and one categorical road-type variable. Missing values are imputed using column-wise means, and class imbalance is addressed through specifying strategies (e.g., SMOTE + random undersampling) for reproducibility. A stratified data split is used for evaluation. We train an XGBoost model with the multi-class objective \texttt{multi:softprob}. Given an input vector $\mathbf{x}$, the model outputs logits $\mathbf{z}(\mathbf{x})\in\mathbb{R}^{K}$, which are converted into probabilities using softmax:
\begin{equation}
\hat{p}_k(\mathbf{x})=\frac{\exp\!\big(z_k(\mathbf{x})\big)}{\sum_{c=1}^{K}\exp\!\big(z_c(\mathbf{x})\big)},\qquad k=1,\ldots,K.
\end{equation}
The probability vector $\hat{\mathbf{p}}(\mathbf{x})$ supports both multi-class prediction and downstream interpretability analysis.

For interpretability, we apply TreeSHAP from the SHAP framework \cite{NIPS2017_8a20a862}, which provides locally faithful, additively decomposed attributions grounded in cooperative game theory. Each prediction is represented as:
\begin{equation}
f(\mathbf{x})=\phi_0+\sum_{j=1}^{M}\phi_j,
\end{equation}
where $\phi_0$ is the baseline output and $\phi_j$ denotes the contribution of feature $j$ to the specific prediction. TreeSHAP yields exact, polynomial-time attributions for tree ensembles. The sign of $\phi_j$ indicates whether a feature increases or decreases the predicted risk, while the magnitude reflects its relative contribution strength.

\begin{figure}[!t]
    \setlength{\abovecaptionskip}{0.1cm}
    \setlength{\belowcaptionskip}{-0.1cm}
    \centering
    \includegraphics[width=0.95\linewidth]{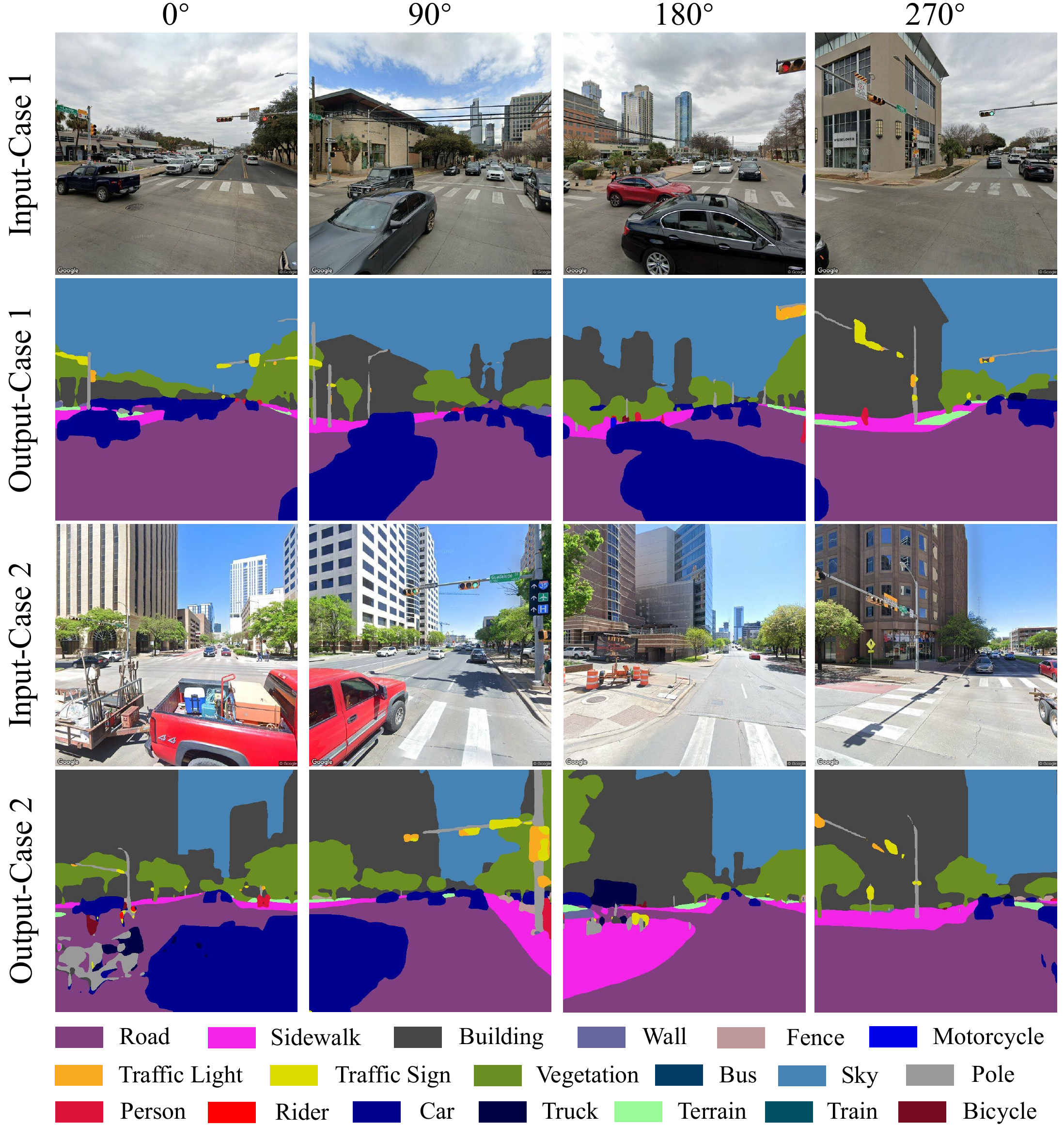}
    \caption{Illustration of input SVIs and corresponding zero-shot semantic segmentation outputs.}
    \label{fig: result}
    \vspace{-0.3cm}
\end{figure}

SHAP values are computed on a held-out test set using a representative background subset from the training distribution to stabilize expectations. We present two complementary perspectives: (1) The global view aggregates sample-level attributions to rank features by mean absolute contribution and visualize overall effect trends. (2) The class-specific view aggregates SHAP values to produce per-class importance and summary plots, enabling direct comparison of key drivers across accident types.

\begin{table}[!t]
    \setlength{\abovecaptionskip}{0.1cm}
    \setlength{\belowcaptionskip}{-0.1cm}
    \centering
    \caption{Generalized Propensity Score (GPS)  model performance for streetscape indicators.}
    \begin{tabular}{lccc}
    \toprule
    Indicators & R$^{2}$ & RMSE & SMD Improvement \\
    \midrule
    BC & 0.841 & 0.012 & 0.077 \\
    SOR & 0.878 & 0.024 & -0.013 \\
    BOR & 0.949 & 0.011 & -0.038 \\
    VOD & 0.711 & 0.715 & 0.119 \\
    VO & 0.953 & 0.021 & 0.046 \\
    DAR & 0.866 & 0.024 & 0.082 \\
    ES & 0.991 & 0.006 & 0.011 \\
    SR & 0.854 & 0.009 & 0.064 \\
    VC & 0.983 & 0.013 & 0.022 \\
    TSI & 0.540 & 0.153 & 0.188 \\
    VD & 0.723 & 0.012 & 0.117 \\
    \bottomrule
    \end{tabular}\label{tab:gps_continuous_performance}
\end{table}

\begin{figure*}[!t]
    \setlength{\abovecaptionskip}{0.1cm}
    \setlength{\belowcaptionskip}{-0.2cm}
    \centering
    \includegraphics[width=0.9\textwidth]{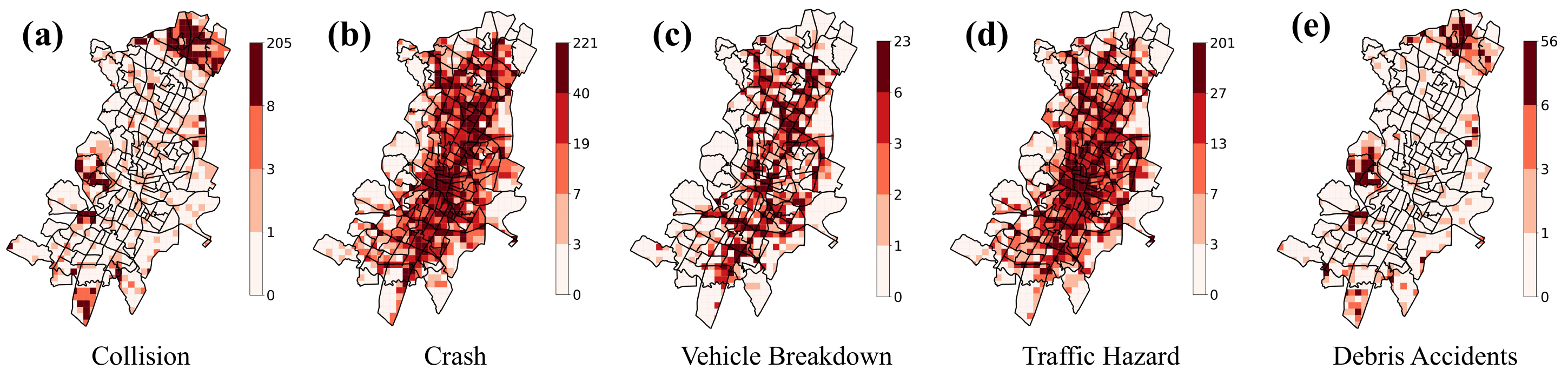}
    \caption{Spatial distribution of five accident categories across the study area.}
    \label{fig: accidenttype}
\end{figure*}

\subsection{Causal Inference Analysis}

We estimate the causal effect of each streetscape feature on accident risk using a two-step procedure. We first apply GPS to balance covariates, followed by a weighted logistic regression to estimate the ATE. Each accident class is formulated as a one-vs-rest binary outcome $Y \in \{0, 1\}$, where a single feature $Z$ is treated as the "treatment" variable and the remaining variable $\mathbf{X}$ are considered as covariates.

To construct GPS, we model the treatment variable $Z$ conditional on covariates $\mathbf{X}$, and derive inverse-probability weights. For categorical features (e.g., road type), we train an XGBoost classifier to estimate class probabilities $\pi_c(\mathbf{x})$, assigning each observation a weight $w_i = 1 / \pi_{z_i}(\mathbf{x}_i)$. For continuous features (e.g., semantic indicators), an XGBoost regressor predicts $\hat{Z}(\mathbf{x})$ , and the weight is derived from the conditional density of the residual $e_i = z_i - \hat{Z}(\mathbf{x}_i)$ under a Gaussian approximation. To ensure numerical stability, weights are truncated at high percentiles. Covariate balance is then evaluated using standardized mean differences (SMD), where for continuous $Z$, we split samples at the median.

%GPS effectively reduces selection bias, as shown by SMD improvements in Tab.~\ref{tab:gps_continuous_performance}, with positive values indicating better balance across most street-view features. \textcolor{red}{High R$^{2}$ and low RMSE further confirm the strong predictive performance of GPS models, ensuring reliable weights for causal inference.}

\begin{figure*}[!t]
    \setlength{\abovecaptionskip}{0.1cm}
    \setlength{\belowcaptionskip}{-0.2cm}
    \centering
    \includegraphics[width=0.95\textwidth]{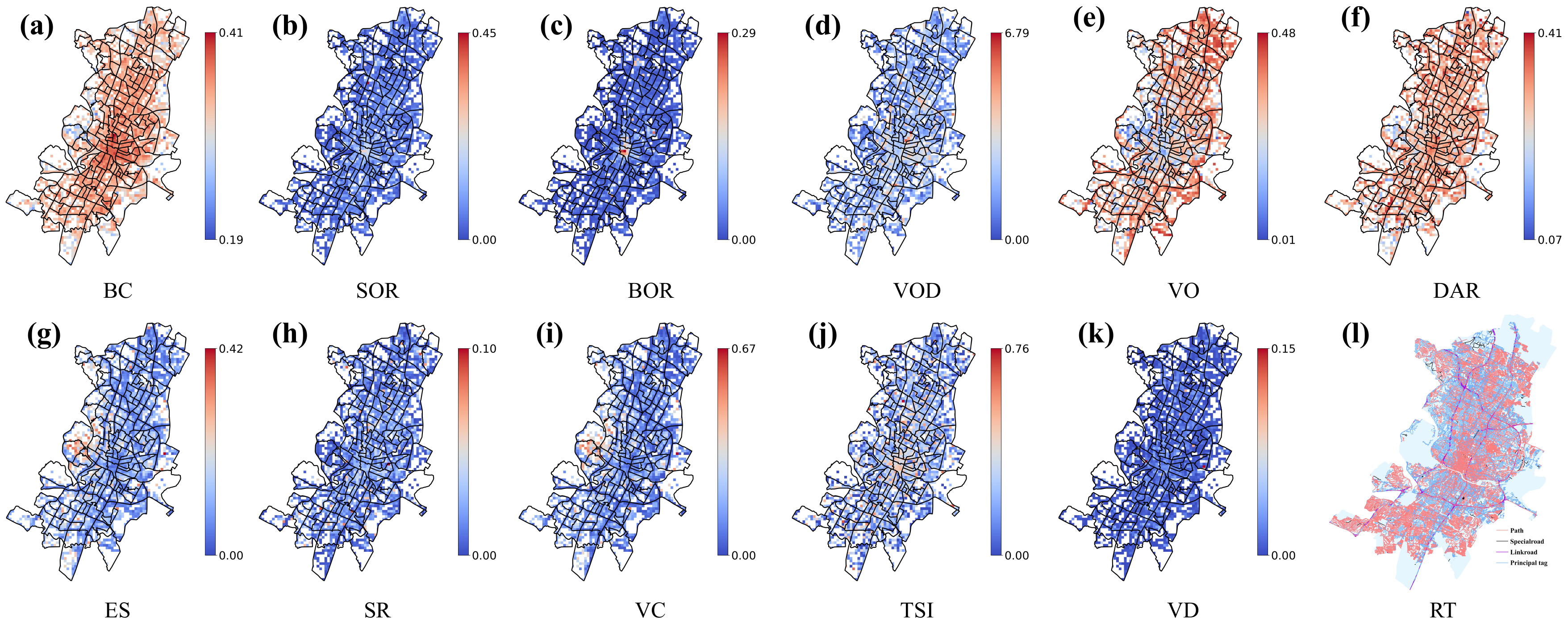}
    \caption{Spatial distribution of street‑view indicators and road types across the study area.}
    \label{fig: fishnet}
\end{figure*}

On the GPS-weighted data, we estimate causal effects using weighted logistic regression. For categorical features $Z$, we include indicator variables for all non-baseline levels. The odds ratios $\mathrm{OR}_c = \exp(\alpha_c)$ quantify the effect of each category relative to the most frequent baseline. For continuous standardized features $Z$, we fit:
\begin{equation}
\Pr(Y = 1 \mid Z) = \mathrm{logit}^{-1}(\beta_0 + \beta_1 Z),
\end{equation}
where the odds ratio is given by $\mathrm{OR} = \exp(\beta_1)$. We compute 95\% confidence intervals and $p$-values using bootstrap standard errors. An $\mathrm{OR} > 1$ indicates a risk-increasing effect, while $\mathrm{OR} < 1$ indicates a risk-reducing effect. This methodology provides a robust foundation for quantifying the causal roles of streetscape features in traffic accidents and supports downstream empirical interpretation.
\section{Results}
\subsection{From SVI Semantic Segmentation to Streetscape Indicator Construction}

We extract structured scene information from street-view images using zero-shot semantic segmentation. As illustrated in Fig.~\ref{fig: result}, each image is segmented into 19 semantic categories encompassing roadway surfaces, pedestrian infrastructure, built structures, vegetation, sky/terrain, vehicles, traffic signs, and common roadside obstructions. These pixel-level masks provide a consistent spatial representation of the urban streetscape across the study area and serve as the foundation for quantitative indicator construction.

To ensure stable representation, each indicator was aggregated across four cardinal views per accident point. This procedure reduced local noise and yielded consistent block-level metrics. The resulting feature set provides interpretable and spatially coherent descriptions of the urban streetscape, bridging raw image content with structured accident analysis. These indicators form the empirical foundation for the subsequent predictive modeling, SHAP interpretability, and causal inference stages of our framework.

\begin{figure}[!t]
    \setlength{\abovecaptionskip}{0.1cm}
    \setlength{\belowcaptionskip}{-0.2cm}
    \centering
    \includegraphics[width=0.95\linewidth]{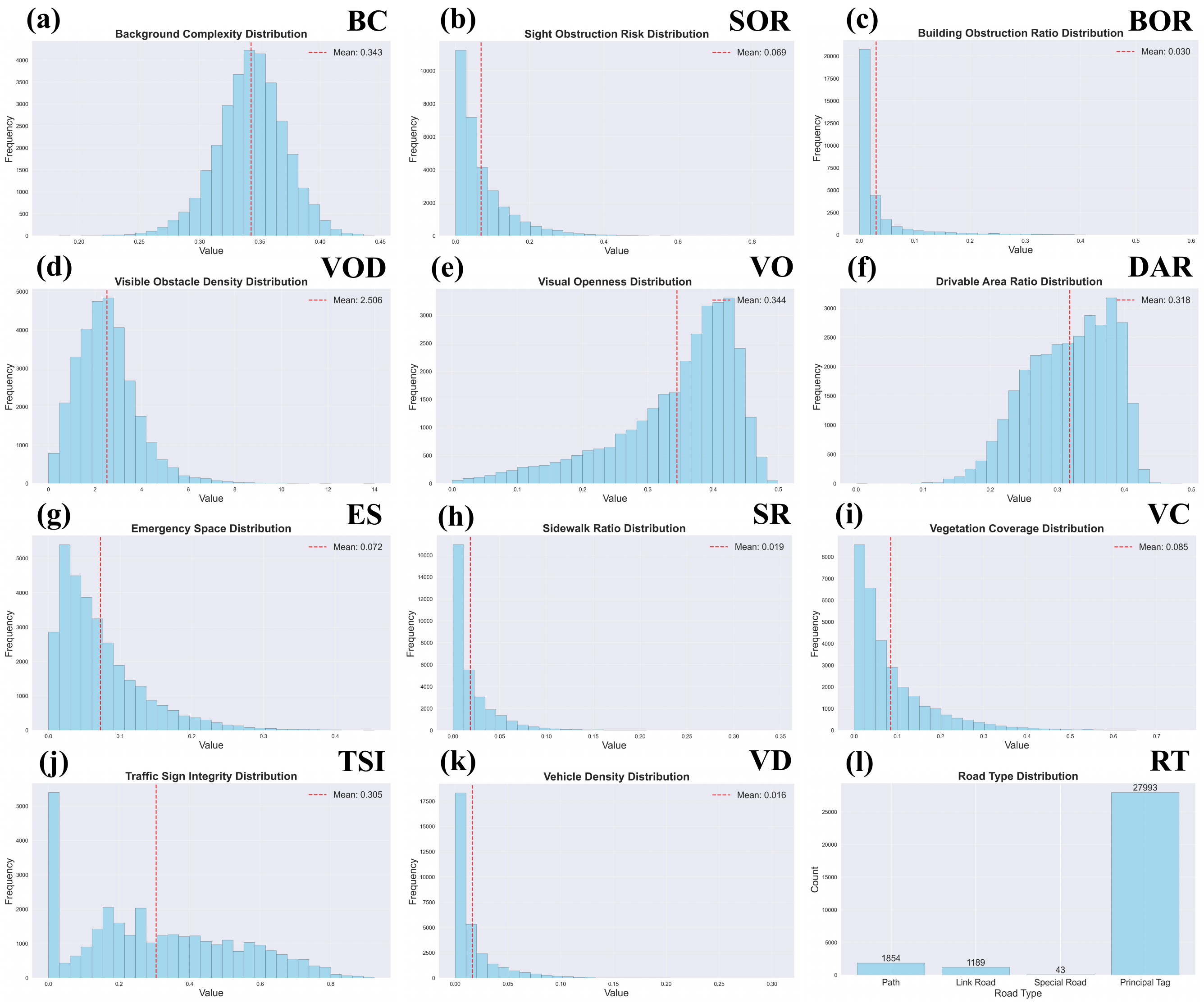}
    \caption{Distribution patterns of street view indicators and road types. Panels (a-k) show histograms of 11 street view indicators with mean values (red dashed lines), while panel (l) displays road type frequency distribution. }
    \label{fig: distribution}
    \vspace{-0.3cm}
\end{figure}

As shown in Fig.~\ref{fig: distribution}, the distribution patterns of street view indicators and road types reveal distinct environmental characteristics across the study area.  Most indicators exhibit left-skewed distributions, indicating generally low environmental risk levels across accident locations. The Visible Obstacle Density (panel d) shows a right-skewed distribution with a mean of 2.506, suggesting varying levels of roadside obstruction across locations. The road network analysis reveals four distinct categories: Principal Tag roads dominate the network , followed by Path roads, Link roads, and Special roads. This hierarchical structure reflects the urban transportation system's organization and provides essential context for understanding how different road types influence accident patterns and environmental risk factors.

These quantitative indicators transform qualitative street-view observations into structured, analyzable features that enable systematic investigation of the relationship between urban environment characteristics and traffic safety outcomes. The comprehensive coverage of both built environment and natural features ensures that our analysis captures the full spectrum of factors that may influence accident occurrence and severity.

\subsection{Spatial Distribution of Accident Types and Street‑view Indicators}

As shown in Fig.~\ref{fig: accidenttype}, accident occurrences cluster along primary road corridors, with high-density hotspots near major arterials and intersections, while peripheral local roads and open-space areas remain low in frequency. Distinct spatial patterns emerge: Collision and Crash cases intensify along central corridors. Vehicle Breakdown aligns with nodal bottlenecks. Traffic Hazards form linear belts along key axes, and Debris Accidents surface as sparse hotspots near high-traffic nodes. These corridor-centric concentrations and peripheral lows reveal a strong spatial coupling with road hierarchy, traffic intensity, and urban activity centers—visually affirming the spatial heterogeneity of risk.

Moreover, we visualize the spatial distribution of accident density and the eleven streetscape indicators using a uniform fishnet grid (Fig.~\ref{fig: fishnet}). Accident occurrences form a corridor-like pattern concentrated along major roads, with several high-density clusters near arterials and intersections—mirroring areas of intense mobility and activity. In contrast, grids with lower accident counts are primarily located in peripheral residential streets and open or low-access zones.

Background Complexity, Visual Openness, and Drivable Area Ratio exhibit clear core–periphery gradients, with higher values in central urban corridors and lower values in peripheral residential areas. Sight Obstruction Risk and Visible Obstacle Density are concentrated near areas with dense street furniture, construction zones, or commercial strips—reflecting greater visual clutter. Emergency Space and Sidewalk Ratio highlight segments with better lateral clearance or pedestrian infrastructure, typically located in planned improvement corridors. In contrast, Building Obstruction Ratio and Vegetation Coverage peak in suburban or green buffer areas, indicating enclosed urban canyons or natural boundaries. Traffic Sign Integrity outlines primary thoroughfares, while Vehicle Density marks traffic-intensive nodes.

These maps show that indicators related to scene complexity (Background Complexity), exposure (Visual Openness), and roadway geometry (Drivable Area Ratio, Emergency Space, Sidewalk Ratio) align closely with high-accident corridors, while enclosure (Building Obstruction Ratio) and greening (Vegetation Coverage) follow complementary, peripheral distributions. This descriptive insight lays the foundation for the following SHAP-based interpretation and causal analysis.

\begin{figure*}[!t]
    \setlength{\abovecaptionskip}{0.1cm}
    \setlength{\belowcaptionskip}{-0.2cm}
    \centering
    \includegraphics[width=0.95\textwidth]{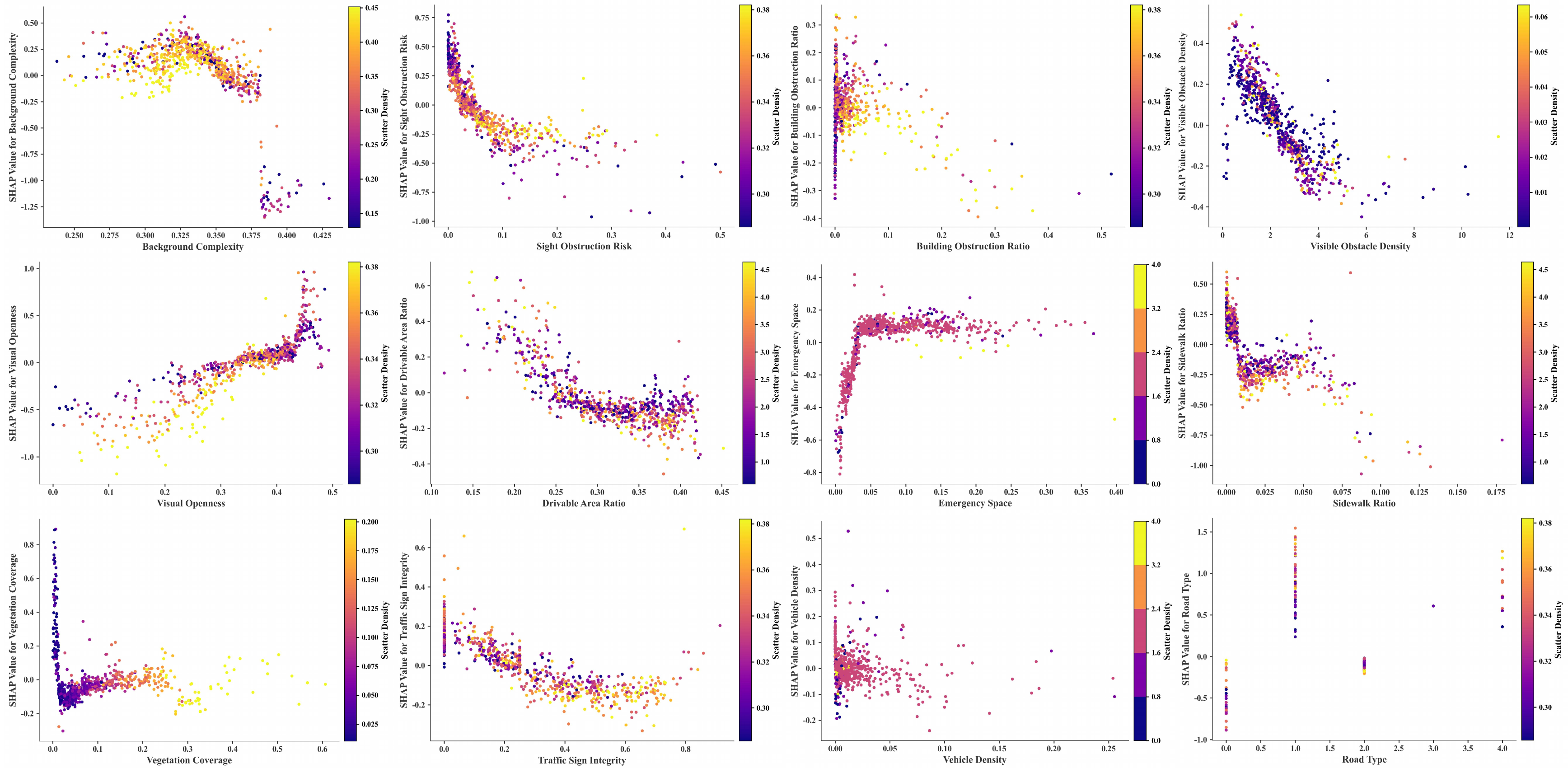}
    \caption{SHAP dependence plots for streetscape indicators.}
    \label{fig: shapdependence}
\end{figure*}

\subsection{Response of Accident Risk to Streetscape Indicators}

We use SHAP analysis to interpret the XGBoost model and quantify the impact of streetscape indicators on accident risk predictions. As shown in the SHAP dependence plots (Fig.~\ref{fig: shapdependence}), the relationships between features and risk are often non-linear. For example, Background Complexity exhibits a clear threshold effect: moderate values (0.30–0.35) are associated with increased predicted risk, while very high values (>0.375) lead to risk reduction. This pattern suggests that moderate visual complexity may reflect busy urban areas with higher accident potential, whereas excessive complexity could trigger more cautious driving behaviors, reducing risk.

Visual Openness shows a strong positive association with accident risk. SHAP values rise monotonically from negative to positive as Visual Openness increases from 0 to 0.5. This counterintuitive trend suggests that more open and unobstructed environments may lead to faster driving or reduced driver vigilance, thereby elevating accident likelihood. In contrast, Sight Obstruction Risk exhibits a negative relationship with predicted accident risk—higher obstruction levels are linked to lower SHAP values. This may indicate that reduced visibility prompts drivers to slow down or become more cautious, effectively lowering the risk of accidents.

Several indicators exhibit threshold-like effects that may inform traffic safety interventions. For example, Emergency Space shows a positive association with accident risk up to around 0.05, beyond which the marginal benefit diminishes. Sidewalk Ratio displays a non-monotonic trend—very low values (<0.025) are linked to elevated risk, while higher ratios contribute to risk reduction, suggesting the existence of optimal sidewalk coverage levels for enhancing pedestrian safety.

The spatial distribution of SHAP values reveals clear heterogeneity in feature effects across urban contexts. High values exhibit stronger feature impacts, while low value regions tend to show weaker effects. This spatial variation underscores the need to account for local urban characteristics when interpreting streetscape indicators for traffic safety assessments.

\begin{figure*}[!t]
    \setlength{\abovecaptionskip}{0.1cm}
    \setlength{\belowcaptionskip}{-0.3cm}
    \centering
    \includegraphics[width=0.95\textwidth]{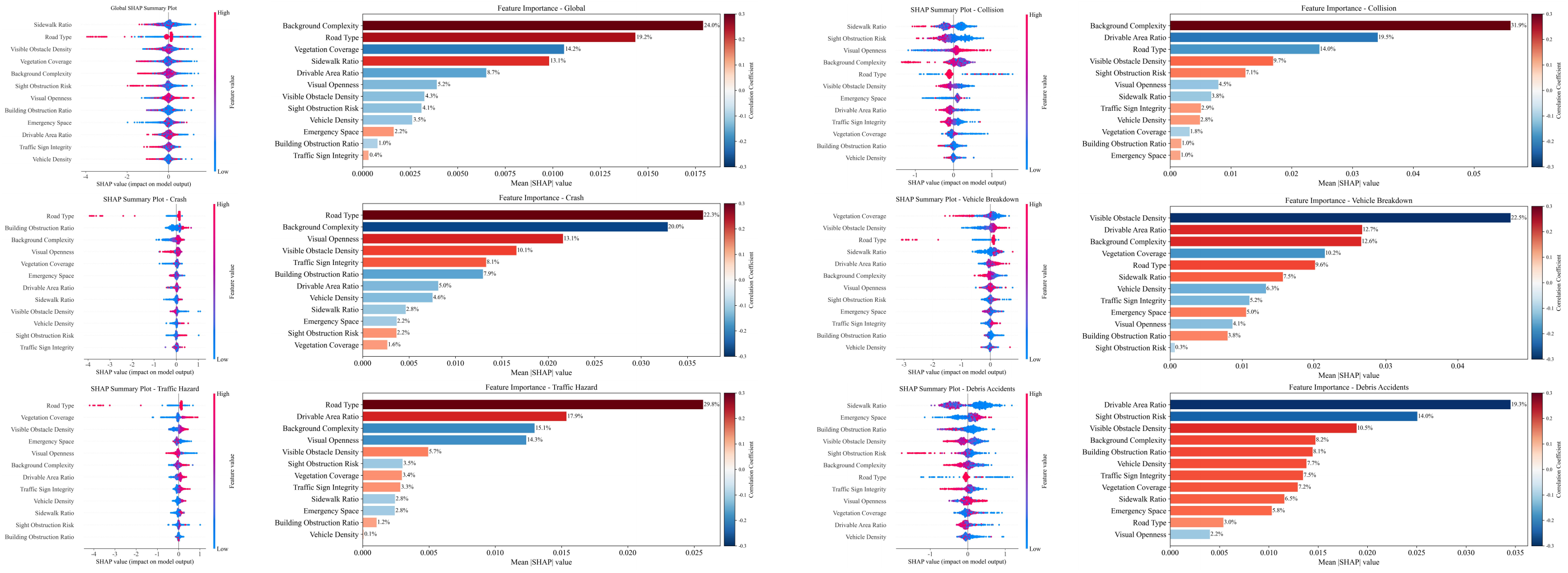}
    \caption{SHAP summary of street‑view indicators for the XGBoost accident‑risk model.}
    \label{fig: shapvalue}
\end{figure*}

As shown in Fig.~\ref{fig: shapvalue}, SHAP summary plots for all features reveal not only their importance but also the direction and distribution of influence. For instance, higher values of Visual Openness tend to increase accident risk (positive SHAP values), while high Background Complexity shows a bimodal effect. The global SHAP analysis shows that Background Complexity is the most influential feature, contributing 24.0\% to the model’s decisions, followed by Road Type at 19.2\% and Vegetation Coverage at 14.2\%. These three indicators account for over 57\% of the total feature importance, highlighting their critical role in traffic accident risk prediction.

Class-specific SHAP analyses reveal distinct patterns across accident types. For Collision accidents, Background Complexity is the most influential feature (31.9\%), followed by Drivable Area Ratio (19.5\%) and Road Type (14.0\%). In Crash cases, Road Type dominates (22.3\%), with Background Complexity contributing (20.0\%). For Vehicle Breakdowns, Visible Obstacle Density (22.5\%) and Drivable Area Ratio (12.7\%) play leading roles, highlighting the impact of visual clutter and road layout on mechanical failures. Both Traffic Hazards and Debris Accidents exhibit similar patterns, where Road Type and Drivable Area Ratio emerge as primary predictors.

The analysis shows that streetscape indicators reflect both physical constraints and perceptual cues that influence driver behavior and accident risk. In particular, the strong predictive power of Background Complexity suggests it may serve as a proxy for urban activity intensity and traffic density, making it a valuable feature for risk assessment in SVI-based safety studies.

\subsection{Causal Effects of Streetscape Indicators on Accident Risk}

Estimating causal effects allows us to move beyond predictive correlations and quantify how specific changes in streetscape indicators directly influence accident risk. Unlike associative models, causal inference explicitly accounts for confounding factors, providing interpretable and actionable insights into how urban form elements affect traffic safety outcomes.

Tab.~\ref{tab:gps_continuous_performance} presents the performance of the GPS models for each streetscape indicator, evaluated by the coefficient of determination ($R^2$), root mean square error (RMSE), and standardized mean difference (SMD) improvement. Overall, most indicators exhibit high $R^2$ values (e.g., ES: 0.991, VC: 0.983), indicating strong predictive ability in modeling treatment assignment based on covariates. RMSE values remain low across most indicators, suggesting reliable model fit. Notably, the majority of indicators achieve positive SMD improvement, confirming the effectiveness of GPS in reducing covariate imbalance and mitigating confounding bias. For example, DAR and VD exhibit both high $R^2$ and significant SMD gains, while some indicators (e.g., SOR and BOR) show minor or negative SMD changes, which may reflect greater estimation variance or treatment overlap challenges. Despite relatively lower performance on discrete variables like TSI, the results overall validate the GPS model's capacity to support robust causal inference in our framework.

\begin{figure*}[!t]
    \setlength{\abovecaptionskip}{0.1cm}
    \setlength{\belowcaptionskip}{-0.1cm}
    \centering
    \includegraphics[width=0.95\textwidth]{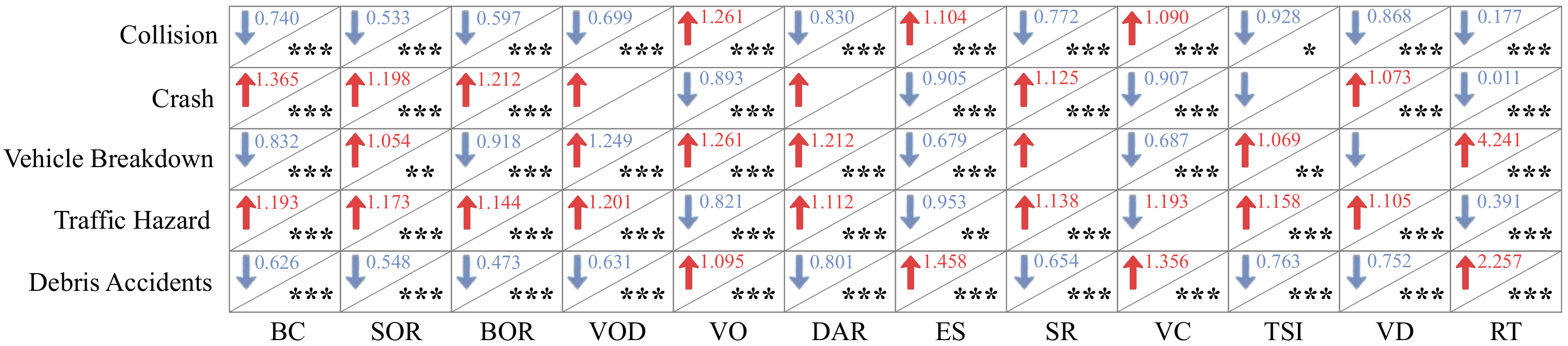}
    \caption{Causal effect matrix of streetscape indicators on different accident types. The numbers in each cell represent the estimated OR (OR = 1.0 indicates no causal effect, OR >
    1.0 suggests increased accident risk, and OR < 1.0 indicates reduced risk) of the corresponding indicator, with red upward arrows indicating increased risk and blue downward arrows indicating reduced risk. Asterisks denote significance levels (* $p < 0.05$, ** $p < 0.01$, *** $p < 0.001$).}
    \label{fig: upanddown}
\end{figure*}

As illustrated in Fig.~\ref{fig: upanddown}, the causal effect matrix reveals distinct patterns in how streetscape indicators influence different accident types after adjusting for confounders via GPS weighting. Each cell shows an odds ratio (OR): OR = 1.0 indicates no causal effect, OR > 1.0 suggests increased accident risk, and OR < 1.0 indicates reduced risk. For instance, OR = 2.0 implies a 100\% increase in risk, while OR = 0.5 implies a 50\% reduction. Notably, Background Complexity exhibits highly heterogeneous effects across accident types. It significantly reduces Collision risk (OR = 0.740, -26\%) but increases Crash risk (OR = 1.365, +36.5\%). This suggests that visual complexity may have context-specific behavioral implications, promoting caution in collision-prone situations but increasing risk in crash-prone environments.

Sight Obstruction Risk consistently exhibits positive causal effects across multiple accident types, with the strongest impacts observed in Vehicle Breakdown and Traffic Hazard categories. This aligns with the expectation that reduced visibility increases accident risk. In contrast, Drivable Area Ratio shows predominantly negative causal effects, particularly pronounced for Debris Accidents, indicating that more extensive drivable space mitigates risk by allowing greater maneuverability. Road Type also emerges as a key factor, showing significant positive causal effects for both Debris Accidents and Crash, suggesting that specific road configurations inherently elevate accident risk.

The analysis confirms that streetscape characteristics act as genuine causal drivers of accident risk, rather than being mere correlates. Over 90\% of feature–accident pairs exhibit statistically significant effects ($p < 0.05$), validating the use of streetscape indicators for traffic safety assessment. This also demonstrates the effectiveness of semantic segmentation in capturing urban morphological features that causally impact accident patterns. Moreover, the heterogeneity of causal effects across accident types underscores the complexity of traffic risk dynamics and highlights the need to consider both feature-specific and accident-type-specific relationships in comprehensive safety evaluations.

\section{Discussion}

\subsection{Impacts of Street‑viewed Indicators on Urban Road Safety}

Our findings consistently demonstrate that streetscape indicators derived from semantic segmentation are not just correlational, but act as causal drivers of accident risk. Both SHAP explanations from the XGBoost classifier and ATE estimates converge on key patterns: 
\begin{itemize}
    \item \textbf{Scene complexity, exposure, and road geometry are the most influential factors.} Specifically, Background Complexity, Visual Openness, and Drivable Area Ratio dominate global importance rankings and exhibit clear monotonic trends in dependence plots. Causal inference further confirms that larger drivable areas and sufficient emergency space reduce accident risk, while excessive visual openness may increase risk—likely due to higher driving speeds and reduced driver vigilance.
    \item \textbf{Visibility-related indicators exhibit context-specific effects.} Both Sight Obstruction Risk and Visible Obstacle Density show positive causal effects for several accident types. Particularly, Vehicle Breakdown and Traffic Hazard support the view that cluttered environments requiring rapid driver responses elevate risk.
    \item \textbf{Pedestrian infrastructure and vegetation play nuanced roles.} Sidewalk Ratio and Vegetation Coverage exhibit mixed but interpretable effects. Sidewalks help reduce risk in pedestrian-exposed corridors, while vegetation is beneficial when it acts as a buffer rather than a visibility obstruction.
    \item \textbf{Road Type remains a key structural determinant of accident risk.} Its strong causal effects for Crash and Debris Accidents suggest that functional classifications and design standards carry inherent risk implications beyond what micro-scale scene features can capture.
\end{itemize}
Collectively, these findings support a two-level safety mechanism: road-level structures (e.g., RT, DAR, ES) define the baseline operating environment, while micro-scene cues (e.g., BC, VO, SOR, VOD, TSI) dynamically modulate driver behavior within environment.

\subsection{Heterogeneity and Complementarity Across Accident Types}

Beyond global effects, class-specific analyses uncover clear patterns of heterogeneity and complementarity. Different accident types are driven by distinct features: Vehicle Breakdown and Traffic Hazard are most sensitive to obstruction-related cues (e.g., SOR, VOD), while Crash and Debris Accidents respond more strongly to road network structure (RT) and maneuvering space (DAR, ES).

Threshold effects observed in the SHAP dependence plots suggest possible regime shifts. For instance, Visual Openness (VO) is positively associated with accident risk up to a certain point, likely due to increased driving speed or reduced attention—beyond which the effect plateaus. Similarly, Emergency Space (ES) shows diminishing returns once a minimal safety width is reached.

Feature combinations often form complementary patterns that align with typical road corridor types. For example, high-risk arterials frequently exhibit large exposure (high VO), limited lateral refuge (low ES), and elevated visual complexity (high BC and VOD). In contrast, lower-risk residential grids tend to balance moderate exposure with more greenery and sidewalk infrastructure, contributing to a safer traffic environment.

The spatial distribution of these feature bundles further supports the plausibility of the proposed mechanisms. Areas with clustered indicator profiles, such as high-exposure arterial corridors or obstruction-dense intersections, consistently overlap with observed accident hotspots. Moreover, the estimated causal directions align with intuitive behavioral logic (e.g., limited visibility increasing hazard likelihood), providing converging evidence for the validity of the causal interpretations.

\subsection{Policy Context and Planning Implications}

Austin has long emphasized data-driven strategies in its transportation safety initiatives. Notably, the Austin Strategic Mobility Plan (ASMP)\footnote{https://www.austintexas.gov/department/austin-strategic-mobility-plan} and Vision Zero Action Plan\footnote{https://www.visionzeroatx.org/reports-data/} highlight the need to reduce fatalities and serious injuries through infrastructure redesign, speed management, and proactive identification of high-risk corridors. Our proposed Semantic4Safety framework aligns closely with these goals by offering a scalable and interpretable approach to assess street-level risk using publicly available imagery and machine learning techniques.

By identifying streetscape features that are causally linked to specific accident types, our framework enables a deeper understanding of how environmental design influences road safety. For example, the finding that sidewalk ratio and vegetation coverage exhibit accident-type-specific effects provides actionable guidance for pedestrian zone retrofitting and green buffer planning, interventions prioritized in Austin's local mobility funding programs.

\subsection{Limitations}

\textit{Temporal alignment remains a key limitation.} In this study, street-view imagery and accident records are not strictly time-aligned. Delays in image acquisition, as well as seasonal and weather-related variations, may influence scene attributes such as visibility or vegetation coverage. Future research should harmonize SVI and crash data at quarterly or annual intervals, and incorporate meteorological and time-of-day covariates to support stratified analyses that account for seasonal confounding and exposure variability.

\textit{Coarseness of Road Type and Design Attributes.} The current road-type variable is label-encoded and lacks fine-grained design information such as speed limits, lane width, channelization, roadside infrastructure, and intersection geometry. Future work should incorporate detailed roadway geometry and asset inventories (e.g., number and width of lanes, median types, access controls, and intersection forms) to improve the explanatory power of structural risk and better isolate design-related causal effects.
\section{Conclusion}

We propose Semantic4Safety, a novel framework that connects street-view imagery (SVI) to urban road safety through semantic segmentation, predictive modeling, and causal inference. Using zero-shot semantic segmentation, we extract 11 scene-level indicators from multiview SVI, map their spatial distributions with road types, and evaluate their relationships with multi-class accident outcomes through an XGBoost classifier, SHAP-based interpretation, and GPS/ATE-based causal estimation. The main conclusions are summarized as follows:
\begin{itemize}
    \item \textbf{Street-view indicators significantly influence accident risk.} Features capturing scene complexity, exposure, and geometry (e.g., BC, VO, DAR, ES) are the most influential. Larger DAR and ES reduce risk, while higher VO tends to increase it.
    \item \textbf{Threshold and spatial effects reveal clear intervention cues.} Several indicators exhibit threshold behaviors in SHAP plots (e.g., VO increases risk up to a point; ES offers diminishing returns). High-risk grids cluster along dense road corridors, while areas with greenery or open buffers show lower risk.
    \item \textbf{Feature importance varies by accident type.} Obstruction-related indicators (SOR, VOD) dominate Vehicle Breakdown and Traffic Hazard, while structural and spatial features (RT, DAR, ES) are key for Crash and Debris Accidents. This heterogeneity supports type-specific, corridor-based safety diagnostics.
\end{itemize}

Overall, Semantic4Safety demonstrates that fine-grained urban morphology extracted from street-view imagery offers actionable and causally meaningful insights for road safety analysis. By integrating spatial interpretation, model explainability, and causal inference, the framework goes beyond correlation-based mapping and provides a rigorous foundation for location-specific safety diagnostics. Future work will aim to improve temporal alignment between SVI and crash records, enrich roadway design features, and explore feature interactions and robustness to strengthen the causal chain from urban context to accident risk.

\section*{Acknowledgments}

This work was supported by the National Natural Science Foundation of China under Project 42371343, Guangdong Basic and Applied Basic Research Foundation with Grant No.2024A1515010986, Program of China Scholarship Council with Grant No.202506380088, the 2025 University of Glasgow Early Career Reward for Excellence on "\emph{Developing Vision–Language Models Enhanced by Geospatial Intelligence}" and the 2025 University of Glasgow Early Career Mobility Scheme (ECMS) Award.

%%
%% The next two lines define the bibliography style to be used, and
%% the bibliography file.
\bibliographystyle{ACM-Reference-Format}
\bibliography{sample-base}

%%% -*-BibTeX-*-
%%% Do NOT edit. File created by BibTeX with style
%%% ACM-Reference-Format-Journals [18-Jan-2012].

\begin{thebibliography}{42}

%%% ====================================================================
%%% NOTE TO THE USER: you can override these defaults by providing
%%% customized versions of any of these macros before the \bibliography
%%% command.  Each of them MUST provide its own final punctuation,
%%% except for \shownote{} and \showURL{}.  The latter two
%%% do not use final punctuation, in order to avoid confusing it with
%%% the Web address.
%%%
%%% To suppress output of a particular field, define its macro to expand
%%% to an empty string, or better, \unskip, like this:
%%%
%%% \newcommand{\showURL}[1]{\unskip}   % LaTeX syntax
%%%
%%% \def \showURL #1{\unskip}           % plain TeX syntax
%%%
%%% ====================================================================

\ifx \showCODEN    \undefined \def \showCODEN     #1{\unskip}     \fi
\ifx \showISBNx    \undefined \def \showISBNx     #1{\unskip}     \fi
\ifx \showISBNxiii \undefined \def \showISBNxiii  #1{\unskip}     \fi
\ifx \showISSN     \undefined \def \showISSN      #1{\unskip}     \fi
\ifx \showLCCN     \undefined \def \showLCCN      #1{\unskip}     \fi
\ifx \shownote     \undefined \def \shownote      #1{#1}          \fi
\ifx \showarticletitle \undefined \def \showarticletitle #1{#1}   \fi
\ifx \showURL      \undefined \def \showURL       {\relax}        \fi
% The following commands are used for tagged output and should be
% invisible to TeX
\providecommand\bibfield[2]{#2}
\providecommand\bibinfo[2]{#2}
\providecommand\natexlab[1]{#1}
\providecommand\showeprint[2][]{arXiv:#2}

\bibitem[Adams and Bischof(1994)]%
        {295913}
\bibfield{author}{\bibinfo{person}{R. Adams} {and} \bibinfo{person}{L. Bischof}.} \bibinfo{year}{1994}\natexlab{}.
\newblock \showarticletitle{Seeded region growing}.
\newblock \bibinfo{journal}{\emph{IEEE Transactions on Pattern Analysis and Machine Intelligence}} \bibinfo{volume}{16}, \bibinfo{number}{6} (\bibinfo{year}{1994}), \bibinfo{pages}{641--647}.
\newblock
\href{https://doi.org/10.1109/34.295913}{doi:\nolinkurl{10.1109/34.295913}}


\bibitem[Arya et~al\mbox{.}(2021)]%
        {Arya_Maeda_Ghosh_Toshniwal_Mraz_Kashiyama_Sekimoto_2021}
\bibfield{author}{\bibinfo{person}{Deeksha Arya}, \bibinfo{person}{Hiroya Maeda}, \bibinfo{person}{Sanjay~Kumar Ghosh}, \bibinfo{person}{Durga Toshniwal}, \bibinfo{person}{Alexander Mraz}, \bibinfo{person}{Takehiro Kashiyama}, {and} \bibinfo{person}{Yoshihide Sekimoto}.} \bibinfo{year}{2021}\natexlab{}.
\newblock \showarticletitle{Deep learning-based road damage detection and classification for multiple countries}.
\newblock \bibinfo{journal}{\emph{Automation in Construction}} (\bibinfo{date}{Dec} \bibinfo{year}{2021}), \bibinfo{pages}{103935}.
\newblock
\href{https://doi.org/10.1016/j.autcon.2021.103935}{doi:\nolinkurl{10.1016/j.autcon.2021.103935}}


\bibitem[Balaban(2022)]%
        {balaban2022understanding}
\bibfield{author}{\bibinfo{person}{{\"O}zg{\"u}n Balaban}.} \bibinfo{year}{2022}\natexlab{}.
\newblock \showarticletitle{Understanding urban leisure walking behavior: Correlations between neighborhood features and fitness tracking data}.
\newblock In \bibinfo{booktitle}{\emph{Artificial Intelligence in Urban Planning and Design}}. \bibinfo{publisher}{Elsevier}, \bibinfo{pages}{245--261}.
\newblock


\bibitem[Biljecki and Ito(2021)]%
        {biljecki2021street}
\bibfield{author}{\bibinfo{person}{Filip Biljecki} {and} \bibinfo{person}{Koichi Ito}.} \bibinfo{year}{2021}\natexlab{}.
\newblock \showarticletitle{Street view imagery in urban analytics and GIS: A review}.
\newblock \bibinfo{journal}{\emph{Landscape and Urban Planning}}  \bibinfo{volume}{215} (\bibinfo{year}{2021}), \bibinfo{pages}{104217}.
\newblock


\bibitem[Breiman(2001)]%
        {Breiman_2001}
\bibfield{author}{\bibinfo{person}{Leo Breiman}.} \bibinfo{year}{2001}\natexlab{}.
\newblock \showarticletitle{Random Forests}.
\newblock \bibinfo{journal}{\emph{Machine Learning}} (\bibinfo{date}{Jan} \bibinfo{year}{2001}), \bibinfo{pages}{5–32}.
\newblock
\href{https://doi.org/10.1023/a:1010933404324}{doi:\nolinkurl{10.1023/a:1010933404324}}


\bibitem[Cai et~al\mbox{.}(2022)]%
        {cai2022applying}
\bibfield{author}{\bibinfo{person}{Qing Cai}, \bibinfo{person}{Mohamed Abdel-Aty}, \bibinfo{person}{Ou Zheng}, {and} \bibinfo{person}{Yina Wu}.} \bibinfo{year}{2022}\natexlab{}.
\newblock \showarticletitle{Applying machine learning and google street view to explore effects of drivers’ visual environment on traffic safety}.
\newblock \bibinfo{journal}{\emph{Transportation research part C: emerging technologies}}  \bibinfo{volume}{135} (\bibinfo{year}{2022}), \bibinfo{pages}{103541}.
\newblock


\bibitem[Cao et~al\mbox{.}(2018)]%
        {Cao_Zhu_Tu_Li_Cao_Liu_Zhang_Qiu_2018}
\bibfield{author}{\bibinfo{person}{Rui Cao}, \bibinfo{person}{Jiasong Zhu}, \bibinfo{person}{Wei Tu}, \bibinfo{person}{Qingquan Li}, \bibinfo{person}{Jinzhou Cao}, \bibinfo{person}{Bozhi Liu}, \bibinfo{person}{Qian Zhang}, {and} \bibinfo{person}{Guoping Qiu}.} \bibinfo{year}{2018}\natexlab{}.
\newblock \showarticletitle{Integrating Aerial and Street View Images for Urban Land Use Classification}.
\newblock \bibinfo{journal}{\emph{Remote Sensing}} (\bibinfo{date}{Sep} \bibinfo{year}{2018}), \bibinfo{pages}{1553}.
\newblock
\href{https://doi.org/10.3390/rs10101553}{doi:\nolinkurl{10.3390/rs10101553}}


\bibitem[Castro and De~Santos-Berbel(2015)]%
        {castro2015spatial}
\bibfield{author}{\bibinfo{person}{Maria Castro} {and} \bibinfo{person}{C{\'e}sar De~Santos-Berbel}.} \bibinfo{year}{2015}\natexlab{}.
\newblock \showarticletitle{Spatial analysis of geometric design consistency and road sight distance}.
\newblock \bibinfo{journal}{\emph{International Journal of Geographical Information Science}} \bibinfo{volume}{29}, \bibinfo{number}{12} (\bibinfo{year}{2015}), \bibinfo{pages}{2061--2074}.
\newblock


\bibitem[Chen and Biljecki(2023)]%
        {CHEN2023104329}
\bibfield{author}{\bibinfo{person}{Shuting Chen} {and} \bibinfo{person}{Filip Biljecki}.} \bibinfo{year}{2023}\natexlab{}.
\newblock \showarticletitle{Automatic assessment of public open spaces using street view imagery}.
\newblock \bibinfo{journal}{\emph{Cities}}  \bibinfo{volume}{137} (\bibinfo{year}{2023}), \bibinfo{pages}{104329}.
\newblock
\showISSN{0264-2751}
\href{https://doi.org/10.1016/j.cities.2023.104329}{doi:\nolinkurl{10.1016/j.cities.2023.104329}}


\bibitem[Chen et~al\mbox{.}(2025a)]%
        {chen2025leveragingdepthlanguageopenvocabulary}
\bibfield{author}{\bibinfo{person}{Siyu Chen}, \bibinfo{person}{Ting Han}, \bibinfo{person}{Chengzheng Fu}, \bibinfo{person}{Changshe Zhang}, \bibinfo{person}{Chaolei Wang}, \bibinfo{person}{Jinhe Su}, \bibinfo{person}{Guorong Cai}, {and} \bibinfo{person}{Meiliu Wu}.} \bibinfo{year}{2025}\natexlab{a}.
\newblock \bibinfo{title}{Leveraging Depth and Language for Open-Vocabulary Domain-Generalized Semantic Segmentation}.
\newblock
\showeprint[arxiv]{2506.09881}~[cs.CV]
\urldef\tempurl%
\url{https://arxiv.org/abs/2506.09881}
\showURL{%
\tempurl}


\bibitem[Chen et~al\mbox{.}(2024)]%
        {chen2024depth}
\bibfield{author}{\bibinfo{person}{Siyu Chen}, \bibinfo{person}{Ting Han}, \bibinfo{person}{Changshe Zhang}, \bibinfo{person}{Weiquan Liu}, \bibinfo{person}{Jinhe Su}, \bibinfo{person}{Zongyue Wang}, {and} \bibinfo{person}{Guorong Cai}.} \bibinfo{year}{2024}\natexlab{}.
\newblock \showarticletitle{Depth Matters: Exploring Deep Interactions of RGB-D for Semantic Segmentation in Traffic Scenes}.
\newblock \bibinfo{journal}{\emph{arXiv preprint arXiv:2409.07995}} (\bibinfo{year}{2024}).
\newblock


\bibitem[Chen et~al\mbox{.}(2025b)]%
        {chen2025hspformer}
\bibfield{author}{\bibinfo{person}{Siyu Chen}, \bibinfo{person}{Ting Han}, \bibinfo{person}{Changshe Zhang}, \bibinfo{person}{Jinhe Su}, \bibinfo{person}{Ruisheng Wang}, \bibinfo{person}{Yiping Chen}, \bibinfo{person}{Zongyue Wang}, {and} \bibinfo{person}{Guorong Cai}.} \bibinfo{year}{2025}\natexlab{b}.
\newblock \showarticletitle{HSPFormer: Hierarchical Spatial Perception Transformer for Semantic Segmentation}.
\newblock \bibinfo{journal}{\emph{IEEE Transactions on Intelligent Transportation Systems}} (\bibinfo{year}{2025}).
\newblock


\bibitem[Cui et~al\mbox{.}(2023)]%
        {CUI2023103537}
\bibfield{author}{\bibinfo{person}{Qinyu Cui}, \bibinfo{person}{Yan Zhang}, \bibinfo{person}{Guang Yang}, \bibinfo{person}{Yiting Huang}, {and} \bibinfo{person}{Yu Chen}.} \bibinfo{year}{2023}\natexlab{}.
\newblock \showarticletitle{Analysing gender differences in the perceived safety from street view imagery}.
\newblock \bibinfo{journal}{\emph{International Journal of Applied Earth Observation and Geoinformation}}  \bibinfo{volume}{124} (\bibinfo{year}{2023}), \bibinfo{pages}{103537}.
\newblock
\showISSN{1569-8432}
\href{https://doi.org/10.1016/j.jag.2023.103537}{doi:\nolinkurl{10.1016/j.jag.2023.103537}}


\bibitem[Dai et~al\mbox{.}(2024)]%
        {Dai02062024}
\bibfield{author}{\bibinfo{person}{Shaoqing Dai}, \bibinfo{person}{Yuchen Li}, \bibinfo{person}{Alfred Stein}, \bibinfo{person}{Shujuan Yang}, {and} \bibinfo{person}{Peng Jia}.} \bibinfo{year}{2024}\natexlab{}.
\newblock \showarticletitle{Street view imagery-based built environment auditing tools: a systematic review}.
\newblock \bibinfo{journal}{\emph{International Journal of Geographical Information Science}} \bibinfo{volume}{38}, \bibinfo{number}{6} (\bibinfo{year}{2024}), \bibinfo{pages}{1136--1157}.
\newblock
\showeprint{https://doi.org/10.1080/13658816.2024.2336034}
\href{https://doi.org/10.1080/13658816.2024.2336034}{doi:\nolinkurl{10.1080/13658816.2024.2336034}}


\bibitem[Davis(2000)]%
        {DAVIS200095}
\bibfield{author}{\bibinfo{person}{Gary~A. Davis}.} \bibinfo{year}{2000}\natexlab{}.
\newblock \showarticletitle{Accident reduction factors and causal inference in traffic safety studies: a review}.
\newblock \bibinfo{journal}{\emph{Accident Analysis \& Prevention}} \bibinfo{volume}{32}, \bibinfo{number}{1} (\bibinfo{year}{2000}), \bibinfo{pages}{95--109}.
\newblock
\showISSN{0001-4575}
\href{https://doi.org/10.1016/S0001-4575(99)00050-0}{doi:\nolinkurl{10.1016/S0001-4575(99)00050-0}}


\bibitem[Fan and Biljecki(2024)]%
        {FAN2024105862}
\bibfield{author}{\bibinfo{person}{Zicheng Fan} {and} \bibinfo{person}{Filip Biljecki}.} \bibinfo{year}{2024}\natexlab{}.
\newblock \showarticletitle{Nighttime Street View Imagery: A new perspective for sensing urban lighting landscape}.
\newblock \bibinfo{journal}{\emph{Sustainable Cities and Society}}  \bibinfo{volume}{116} (\bibinfo{year}{2024}), \bibinfo{pages}{105862}.
\newblock
\showISSN{2210-6707}
\href{https://doi.org/10.1016/j.scs.2024.105862}{doi:\nolinkurl{10.1016/j.scs.2024.105862}}


\bibitem[Fan et~al\mbox{.}({[n.\,d.]})]%
        {Fan_Zhang_Loo_Ratti}
\bibfield{author}{\bibinfo{person}{Zhuangyuan Fan}, \bibinfo{person}{Fan Zhang}, \bibinfo{person}{BeckyPY Loo}, {and} \bibinfo{person}{Carlo Ratti}.} \bibinfo{year}{[n.\,d.]}\natexlab{}.
\newblock \showarticletitle{Urban visual intelligence: Uncovering hidden city profiles with street view images}.
\newblock  (\bibinfo{year}{[n.\,d.]}).
\newblock


\bibitem[Gao et~al\mbox{.}(2021)]%
        {9724728}
\bibfield{author}{\bibinfo{person}{ZhiGang Gao}, \bibinfo{person}{ShengYuan Guan}, {and} \bibinfo{person}{MianSheng Guo}.} \bibinfo{year}{2021}\natexlab{}.
\newblock \showarticletitle{Semantic Segmentation of Street View Image Based on Fully Convolutional Neural Networks}. In \bibinfo{booktitle}{\emph{2021 2nd International Seminar on Artificial Intelligence, Networking and Information Technology (AINIT)}}. \bibinfo{pages}{400--404}.
\newblock
\href{https://doi.org/10.1109/AINIT54228.2021.00084}{doi:\nolinkurl{10.1109/AINIT54228.2021.00084}}


\bibitem[Guo et~al\mbox{.}(2024)]%
        {guo2024fusion}
\bibfield{author}{\bibinfo{person}{Wentong Guo}, \bibinfo{person}{Cheng Xu}, {and} \bibinfo{person}{Sheng Jin}.} \bibinfo{year}{2024}\natexlab{}.
\newblock \showarticletitle{Fusion of satellite and street view data for urban traffic accident hotspot identification}.
\newblock \bibinfo{journal}{\emph{International Journal of Applied Earth Observation and Geoinformation}}  \bibinfo{volume}{130} (\bibinfo{year}{2024}), \bibinfo{pages}{103853}.
\newblock


\bibitem[Han et~al\mbox{.}(2024)]%
        {han2024epurate}
\bibfield{author}{\bibinfo{person}{Ting Han}, \bibinfo{person}{Siyu Chen}, \bibinfo{person}{Chuanmu Li}, \bibinfo{person}{Zongyue Wang}, \bibinfo{person}{Jinhe Su}, \bibinfo{person}{Min Huang}, {and} \bibinfo{person}{Guorong Cai}.} \bibinfo{year}{2024}\natexlab{}.
\newblock \showarticletitle{Epurate-net: Efficient progressive uncertainty refinement analysis for traffic environment urban road detection}.
\newblock \bibinfo{journal}{\emph{IEEE Transactions on Intelligent Transportation Systems}} \bibinfo{volume}{25}, \bibinfo{number}{7} (\bibinfo{year}{2024}), \bibinfo{pages}{6617--6632}.
\newblock


\bibitem[Hartigan and Wong(1979)]%
        {8ddb7f85-9a8c-3829-b04e-0476a67eb0fd}
\bibfield{author}{\bibinfo{person}{J.~A. Hartigan} {and} \bibinfo{person}{M.~A. Wong}.} \bibinfo{year}{1979}\natexlab{}.
\newblock \showarticletitle{Algorithm AS 136: A K-Means Clustering Algorithm}.
\newblock \bibinfo{journal}{\emph{Journal of the Royal Statistical Society. Series C (Applied Statistics)}} \bibinfo{volume}{28}, \bibinfo{number}{1} (\bibinfo{year}{1979}), \bibinfo{pages}{100--108}.
\newblock
\showISSN{00359254, 14679876}
\urldef\tempurl%
\url{http://www.jstor.org/stable/2346830}
\showURL{%
\tempurl}


\bibitem[Hearst et~al\mbox{.}(1998)]%
        {Hearst_Dumais_Osuna_Platt_Scholkopf_1998}
\bibfield{author}{\bibinfo{person}{M.A. Hearst}, \bibinfo{person}{S.T. Dumais}, \bibinfo{person}{E. Osuna}, \bibinfo{person}{J. Platt}, {and} \bibinfo{person}{B. Scholkopf}.} \bibinfo{year}{1998}\natexlab{}.
\newblock \showarticletitle{Support vector machines}.
\newblock \bibinfo{journal}{\emph{IEEE Intelligent Systems and their Applications}} (\bibinfo{date}{Jul} \bibinfo{year}{1998}), \bibinfo{pages}{18–28}.
\newblock
\href{https://doi.org/10.1109/5254.708428}{doi:\nolinkurl{10.1109/5254.708428}}


\bibitem[Hu et~al\mbox{.}(2023)]%
        {hu2023uncovering}
\bibfield{author}{\bibinfo{person}{Sheng Hu}, \bibinfo{person}{Hanfa Xing}, \bibinfo{person}{Wei Luo}, \bibinfo{person}{Liang Wu}, \bibinfo{person}{Yongyang Xu}, \bibinfo{person}{Weiming Huang}, \bibinfo{person}{Wenkai Liu}, {and} \bibinfo{person}{Tianqi Li}.} \bibinfo{year}{2023}\natexlab{}.
\newblock \showarticletitle{Uncovering the association between traffic crashes and street-level built-environment features using street view images}.
\newblock \bibinfo{journal}{\emph{International Journal of Geographical Information Science}} \bibinfo{volume}{37}, \bibinfo{number}{11} (\bibinfo{year}{2023}), \bibinfo{pages}{2367--2391}.
\newblock


\bibitem[Irfan et~al\mbox{.}(2024)]%
        {irfan2024toward}
\bibfield{author}{\bibinfo{person}{Muhammad~Sami Irfan}, \bibinfo{person}{Sagar Dasgupta}, {and} \bibinfo{person}{Mizanur Rahman}.} \bibinfo{year}{2024}\natexlab{}.
\newblock \showarticletitle{Toward transportation digital twin systems for traffic safety and mobility: A review}.
\newblock \bibinfo{journal}{\emph{IEEE Internet of Things Journal}} \bibinfo{volume}{11}, \bibinfo{number}{14} (\bibinfo{year}{2024}), \bibinfo{pages}{24581--24603}.
\newblock


\bibitem[Jiao and Wang(2023)]%
        {Jiao_Wang_2023}
\bibfield{author}{\bibinfo{person}{Junfeng Jiao} {and} \bibinfo{person}{Huihai Wang}.} \bibinfo{year}{2023}\natexlab{}.
\newblock \showarticletitle{Forecasting Traffic Speed during Daytime from Google Street View Images using Deep Learning}.
\newblock  (\bibinfo{date}{May} \bibinfo{year}{2023}).
\newblock


\bibitem[Kang et~al\mbox{.}(2020)]%
        {Kang_Zhang_Gao_Lin_Liu_2020}
\bibfield{author}{\bibinfo{person}{Yuhao Kang}, \bibinfo{person}{Fan Zhang}, \bibinfo{person}{Song Gao}, \bibinfo{person}{Hui Lin}, {and} \bibinfo{person}{Yu Liu}.} \bibinfo{year}{2020}\natexlab{}.
\newblock \showarticletitle{A review of urban physical environment sensing using street view imagery in public health studies}.
\newblock \bibinfo{journal}{\emph{Annals of GIS}} (\bibinfo{date}{Jul} \bibinfo{year}{2020}), \bibinfo{pages}{261–275}.
\newblock
\href{https://doi.org/10.1080/19475683.2020.1791954}{doi:\nolinkurl{10.1080/19475683.2020.1791954}}


\bibitem[Koo et~al\mbox{.}(2022)]%
        {Koo_Guhathakurta_Botchwey_2022}
\bibfield{author}{\bibinfo{person}{Bon~Woo Koo}, \bibinfo{person}{Subhrajit Guhathakurta}, {and} \bibinfo{person}{Nisha Botchwey}.} \bibinfo{year}{2022}\natexlab{}.
\newblock \showarticletitle{How are Neighborhood and Street-Level Walkability Factors Associated with Walking Behaviors? A Big Data Approach Using Street View Images}.
\newblock \bibinfo{journal}{\emph{Environment and Behavior}} (\bibinfo{date}{Jan} \bibinfo{year}{2022}), \bibinfo{pages}{211–241}.
\newblock
\href{https://doi.org/10.1177/00139165211014609}{doi:\nolinkurl{10.1177/00139165211014609}}


\bibitem[Li et~al\mbox{.}(2015)]%
        {Li_Zhang_Li_Ricard_Meng_Zhang_2015}
\bibfield{author}{\bibinfo{person}{Xiaojiang Li}, \bibinfo{person}{Chuanrong Zhang}, \bibinfo{person}{Weidong Li}, \bibinfo{person}{Robert Ricard}, \bibinfo{person}{Qingyan Meng}, {and} \bibinfo{person}{Weixing Zhang}.} \bibinfo{year}{2015}\natexlab{}.
\newblock \showarticletitle{Assessing street-level urban greenery using Google Street View and a modified green view index}.
\newblock \bibinfo{journal}{\emph{Urban Forestry \&amp; Urban Greening}} (\bibinfo{date}{Jan} \bibinfo{year}{2015}), \bibinfo{pages}{675–685}.
\newblock
\href{https://doi.org/10.1016/j.ufug.2015.06.006}{doi:\nolinkurl{10.1016/j.ufug.2015.06.006}}


\bibitem[Liang et~al\mbox{.}(2017)]%
        {Liang_Gong_Sun_Zhou_Li_Li_Liu_Shen_2017}
\bibfield{author}{\bibinfo{person}{Jianming Liang}, \bibinfo{person}{Jianhua Gong}, \bibinfo{person}{Jun Sun}, \bibinfo{person}{Jieping Zhou}, \bibinfo{person}{Wenhang Li}, \bibinfo{person}{Yi Li}, \bibinfo{person}{Jin Liu}, {and} \bibinfo{person}{Shen Shen}.} \bibinfo{year}{2017}\natexlab{}.
\newblock \showarticletitle{Automatic Sky View Factor Estimation from Street View Photographs—A Big Data Approach}.
\newblock \bibinfo{journal}{\emph{Remote Sensing}} (\bibinfo{date}{May} \bibinfo{year}{2017}), \bibinfo{pages}{411}.
\newblock
\href{https://doi.org/10.3390/rs9050411}{doi:\nolinkurl{10.3390/rs9050411}}


\bibitem[Lundberg and Lee(2017)]%
        {NIPS2017_8a20a862}
\bibfield{author}{\bibinfo{person}{Scott~M Lundberg} {and} \bibinfo{person}{Su-In Lee}.} \bibinfo{year}{2017}\natexlab{}.
\newblock \showarticletitle{A Unified Approach to Interpreting Model Predictions}. In \bibinfo{booktitle}{\emph{Advances in Neural Information Processing Systems}}, \bibfield{editor}{\bibinfo{person}{I.~Guyon}, \bibinfo{person}{U.~Von Luxburg}, \bibinfo{person}{S.~Bengio}, \bibinfo{person}{H.~Wallach}, \bibinfo{person}{R.~Fergus}, \bibinfo{person}{S.~Vishwanathan}, {and} \bibinfo{person}{R.~Garnett}} (Eds.), Vol.~\bibinfo{volume}{30}. \bibinfo{publisher}{Curran Associates, Inc.}
\newblock
\urldef\tempurl%
\url{https://proceedings.neurips.cc/paper_files/paper/2017/file/8a20a8621978632d76c43dfd28b67767-Paper.pdf}
\showURL{%
\tempurl}


\bibitem[Ma et~al\mbox{.}(2024)]%
        {Ma01112024}
\bibfield{author}{\bibinfo{person}{Xiaobo Ma}, \bibinfo{person}{Abolfazl Karimpour}, {and} \bibinfo{person}{Yao-Jan Wu}.} \bibinfo{year}{2024}\natexlab{}.
\newblock \showarticletitle{Eliminating the impacts of traffic volume variation on before and after studies: a causal inference approach}.
\newblock \bibinfo{journal}{\emph{Journal of Intelligent Transportation Systems}} \bibinfo{volume}{28}, \bibinfo{number}{6} (\bibinfo{year}{2024}), \bibinfo{pages}{921--935}.
\newblock
\showeprint{https://doi.org/10.1080/15472450.2023.2245327}
\href{https://doi.org/10.1080/15472450.2023.2245327}{doi:\nolinkurl{10.1080/15472450.2023.2245327}}


\bibitem[ROSENBAUM and RUBIN(1983)]%
        {ROSENBAUM_RUBIN_1983}
\bibfield{author}{\bibinfo{person}{PAUL~R. ROSENBAUM} {and} \bibinfo{person}{DONALD~B. RUBIN}.} \bibinfo{year}{1983}\natexlab{}.
\newblock \showarticletitle{The central role of the propensity score in observational studies for causal effects}.
\newblock \bibinfo{journal}{\emph{Biometrika}} (\bibinfo{date}{Jan} \bibinfo{year}{1983}), \bibinfo{pages}{41–55}.
\newblock
\href{https://doi.org/10.1093/biomet/70.1.41}{doi:\nolinkurl{10.1093/biomet/70.1.41}}


\bibitem[Rui(2023)]%
        {RUI2023104472}
\bibfield{author}{\bibinfo{person}{Jin Rui}.} \bibinfo{year}{2023}\natexlab{}.
\newblock \showarticletitle{Measuring streetscape perceptions from driveways and sidewalks to inform pedestrian-oriented street renewal in Düsseldorf}.
\newblock \bibinfo{journal}{\emph{Cities}}  \bibinfo{volume}{141} (\bibinfo{year}{2023}), \bibinfo{pages}{104472}.
\newblock
\showISSN{0264-2751}
\href{https://doi.org/10.1016/j.cities.2023.104472}{doi:\nolinkurl{10.1016/j.cities.2023.104472}}


\bibitem[Rui et~al\mbox{.}(2016)]%
        {rui2016network}
\bibfield{author}{\bibinfo{person}{Yikang Rui}, \bibinfo{person}{Zaigui Yang}, \bibinfo{person}{Tianlu Qian}, \bibinfo{person}{Shoaib Khalid}, \bibinfo{person}{Nan Xia}, {and} \bibinfo{person}{Jiechen Wang}.} \bibinfo{year}{2016}\natexlab{}.
\newblock \showarticletitle{Network-constrained and category-based point pattern analysis for Suguo retail stores in Nanjing, China}.
\newblock \bibinfo{journal}{\emph{International Journal of Geographical Information Science}} \bibinfo{volume}{30}, \bibinfo{number}{2} (\bibinfo{year}{2016}), \bibinfo{pages}{186--199}.
\newblock


\bibitem[Rzotkiewicz et~al\mbox{.}(2018)]%
        {Rzotkiewicz_Pearson_Dougherty_Shortridge_Wilson_2018}
\bibfield{author}{\bibinfo{person}{Amanda Rzotkiewicz}, \bibinfo{person}{Amber~L. Pearson}, \bibinfo{person}{Benjamin~V. Dougherty}, \bibinfo{person}{Ashton Shortridge}, {and} \bibinfo{person}{Nick Wilson}.} \bibinfo{year}{2018}\natexlab{}.
\newblock \showarticletitle{Systematic review of the use of Google Street View in health research: Major themes, strengths, weaknesses and possibilities for future research}.
\newblock \bibinfo{journal}{\emph{Health \&amp; Place}} (\bibinfo{date}{Jul} \bibinfo{year}{2018}), \bibinfo{pages}{240–246}.
\newblock
\href{https://doi.org/10.1016/j.healthplace.2018.07.001}{doi:\nolinkurl{10.1016/j.healthplace.2018.07.001}}


\bibitem[Shu et~al\mbox{.}(2021)]%
        {Shu_Yan_Xu_2021}
\bibfield{author}{\bibinfo{person}{Zekai Shu}, \bibinfo{person}{Zhaoyu Yan}, {and} \bibinfo{person}{Xihang Xu}.} \bibinfo{year}{2021}\natexlab{}.
\newblock \showarticletitle{Pavement Crack Detection Method of Street View Images Based on Deep Learning}.
\newblock \bibinfo{journal}{\emph{Journal of Physics: Conference Series}} \bibinfo{volume}{1952}, \bibinfo{number}{2} (\bibinfo{date}{Jun} \bibinfo{year}{2021}), \bibinfo{pages}{022043}.
\newblock
\href{https://doi.org/10.1088/1742-6596/1952/2/022043}{doi:\nolinkurl{10.1088/1742-6596/1952/2/022043}}


\bibitem[Xiao and Quan(2009)]%
        {5459249}
\bibfield{author}{\bibinfo{person}{Jianxiong Xiao} {and} \bibinfo{person}{Long Quan}.} \bibinfo{year}{2009}\natexlab{}.
\newblock \showarticletitle{Multiple view semantic segmentation for street view images}. In \bibinfo{booktitle}{\emph{2009 IEEE 12th International Conference on Computer Vision}}. \bibinfo{pages}{686--693}.
\newblock
\href{https://doi.org/10.1109/ICCV.2009.5459249}{doi:\nolinkurl{10.1109/ICCV.2009.5459249}}


\bibitem[Yang et~al\mbox{.}(2025)]%
        {yang2025theory}
\bibfield{author}{\bibinfo{person}{Hao Yang}, \bibinfo{person}{X~Angela Yao}, \bibinfo{person}{Farnoosh Roozkhosh}, \bibinfo{person}{Ruowei Liu}, {and} \bibinfo{person}{Gengchen Mai}.} \bibinfo{year}{2025}\natexlab{}.
\newblock \showarticletitle{From theory to deep learning: Understanding the impact of geographic context factors on traffic violations}.
\newblock \bibinfo{journal}{\emph{Computers, Environment and Urban Systems}}  \bibinfo{volume}{119} (\bibinfo{year}{2025}), \bibinfo{pages}{102268}.
\newblock


\bibitem[Yang et~al\mbox{.}(2009)]%
        {Yang_Zhao_Mcbride_Gong_2009}
\bibfield{author}{\bibinfo{person}{Jun Yang}, \bibinfo{person}{Linsen Zhao}, \bibinfo{person}{Joe Mcbride}, {and} \bibinfo{person}{Peng Gong}.} \bibinfo{year}{2009}\natexlab{}.
\newblock \showarticletitle{Can you see green? Assessing the visibility of urban forests in cities}.
\newblock \bibinfo{journal}{\emph{Landscape and Urban Planning}} (\bibinfo{date}{Jun} \bibinfo{year}{2009}), \bibinfo{pages}{97–104}.
\newblock
\href{https://doi.org/10.1016/j.landurbplan.2008.12.004}{doi:\nolinkurl{10.1016/j.landurbplan.2008.12.004}}


\bibitem[Yao et~al\mbox{.}(2021)]%
        {10.1145/3444944}
\bibfield{author}{\bibinfo{person}{Liuyi Yao}, \bibinfo{person}{Zhixuan Chu}, \bibinfo{person}{Sheng Li}, \bibinfo{person}{Yaliang Li}, \bibinfo{person}{Jing Gao}, {and} \bibinfo{person}{Aidong Zhang}.} \bibinfo{year}{2021}\natexlab{}.
\newblock \showarticletitle{A Survey on Causal Inference}.
\newblock \bibinfo{journal}{\emph{ACM Trans. Knowl. Discov. Data}} \bibinfo{volume}{15}, \bibinfo{number}{5}, Article \bibinfo{articleno}{74} (\bibinfo{date}{May} \bibinfo{year}{2021}), \bibinfo{numpages}{46}~pages.
\newblock
\showISSN{1556-4681}
\href{https://doi.org/10.1145/3444944}{doi:\nolinkurl{10.1145/3444944}}


\bibitem[Ye et~al\mbox{.}(2025)]%
        {ye2025street}
\bibfield{author}{\bibinfo{person}{Xinyue Ye}, \bibinfo{person}{Shoujia Li}, \bibinfo{person}{Wenjing Gong}, \bibinfo{person}{Xiao Li}, \bibinfo{person}{Xinyu Li}, \bibinfo{person}{Bahar Dadashova}, \bibinfo{person}{Wei Li}, \bibinfo{person}{Jiaxin Du}, {and} \bibinfo{person}{Dayong Wu}.} \bibinfo{year}{2025}\natexlab{}.
\newblock \showarticletitle{Street View Imagery in Traffic Crash and Road Safety Analysis: A Review}.
\newblock \bibinfo{journal}{\emph{Applied Spatial Analysis and Policy}} \bibinfo{volume}{18}, \bibinfo{number}{2} (\bibinfo{year}{2025}), \bibinfo{pages}{50}.
\newblock


\bibitem[Zhang et~al\mbox{.}(2023)]%
        {ijgi12020073}
\bibfield{author}{\bibinfo{person}{Yunfei Zhang}, \bibinfo{person}{Fangqi Zhu}, \bibinfo{person}{Qiuping Li}, \bibinfo{person}{Zehang Qiu}, {and} \bibinfo{person}{Yajun Xie}.} \bibinfo{year}{2023}\natexlab{}.
\newblock \showarticletitle{Exploring Spatiotemporal Patterns of Expressway Traffic Accidents Based on Density Clustering and Bayesian Network}.
\newblock \bibinfo{journal}{\emph{ISPRS International Journal of Geo-Information}} \bibinfo{volume}{12}, \bibinfo{number}{2} (\bibinfo{year}{2023}).
\newblock
\showISSN{2220-9964}
\href{https://doi.org/10.3390/ijgi12020073}{doi:\nolinkurl{10.3390/ijgi12020073}}


\end{thebibliography}

\end{document}